\PassOptionsToPackage{table}{xcolor}
\documentclass{article}
\usepackage{iclr2027_conference,times}
\iclrfinalcopy

\usepackage{amsmath,amsfonts,bm}

\def\eqref#1{equation~\ref{#1}}

\def\1{\bm{1}}

\DeclareMathAlphabet{\mathsfit}{\encodingdefault}{\sfdefault}{m}{sl}
\SetMathAlphabet{\mathsfit}{bold}{\encodingdefault}{\sfdefault}{bx}{n}

\usepackage{hyperref}
\usepackage{amsmath,amssymb,bm}
\usepackage{graphicx}
\usepackage{microtype}
\usepackage{capt-of}
\usepackage[ruled,vlined,linesnumbered]{algorithm2e}

\usepackage{url}
\usepackage{multirow}
\usepackage{booktabs}
\usepackage[table]{xcolor}
\usepackage{xspace}

\definecolor{duetD}{HTML}{5B4B8A}
\definecolor{duetU}{HTML}{7B6BA8}
\definecolor{duetE}{HTML}{9B8BC6}
\definecolor{duetT}{HTML}{B8AAD9}
\definecolor{duetG}{HTML}{E8956B}
\definecolor{duetE2}{HTML}{EA7A52}
\definecolor{duetN}{HTML}{E65C3B}
\newcommand{\method}{%
  \textcolor{duetD}{D}%
  \textcolor{duetU}{u}%
  \textcolor{duetE}{e}%
  \textcolor{duetT}{t}%
  \textcolor{duetG}{G}%
  \textcolor{duetE2}{e}%
  \textcolor{duetN}{n}\xspace
}
\definecolor{grayrow}{gray}{0.8}
\definecolor{tblBlue}{HTML}{E4EDF9}
\definecolor{tblTeal}{HTML}{E4F2EE}
\definecolor{tblPeach}{HTML}{FBEBD8}
\definecolor{tblLilac}{HTML}{EEE8F7}
\definecolor{tblStripe}{HTML}{F5F7FB}
\definecolor{tblInk}{HTML}{304766}
\definecolor{tblBest}{HTML}{A63D40}
\definecolor{tblSecond}{HTML}{285F93}
\definecolor{tblGood}{HTML}{267565}
\definecolor{tblWarm}{HTML}{956021}
\newcommand{\tblbest}[1]{\textcolor{tblBest}{\textbf{#1}}}
\newcommand{\tblsecond}[1]{\textcolor{tblSecond}{\underline{#1}}}
\newcommand{\layout}{\mathcal{S}}

\title{Planning And Rendering In Concert: Deep-fusion OF AUTOREGRESSIVE LAYOUTS AND DIFFUSION FOR VISUAL TEXT GENERATION}

\author{Guanqiao Chen$^{1}$, Jingru Tan$^{2}$, Dongxing Mao$^{2}$, Catherine Chen$^{3}$\\
\bfseries Zijian Du$^{4}$, Libo Qin$^{2}$, Hu Jian Guo$^{5}$, Alex Jinpeng Wang$^{2}$\\[0.4em]
$^{1}$University of Science and Technology of China\\
$^{2}$Central South University\\
$^{3}$Louisiana State University\\
$^{4}$Arizona State University\\
$^{5}$Sun Yat-sen University}

\begin{document}
\maketitle

\begin{abstract}
Generating text-rich images from prompts requires both textual fidelity and the coherent integration of text into the surrounding image. An explicit layout can provide structured guidance about what text should appear and where, but a well-formed plan alone does not guarantee that the renderer will realize it faithfully. Existing layout-based AR--diffusion systems typically optimize planning and rendering separately, preventing the planner's representations from being adapted jointly with image synthesis.
We introduce \method, an autonomous visual text generator built on DeepFusion, which jointly learns autoregressive planning and continuous diffusion rendering. DeepFusion conditions a diffusion transformer on the planner's prompt and bbox--content hidden states, allowing rendering supervision to shape the representations connecting textual plans with visual outputs. Its joint objective combines autoregressive plan supervision, text-region-weighted diffusion learning, and auxiliary coordinate supervision to maintain structured planning, emphasize text-bearing regions, and improve the spatial precision of planner representations. During inference, Phase-Aware Attention Modulation strengthens the correspondence between image regions and their matched coordinate and content states, facilitating region-specific execution of the generated plan.
With a 2B planner and a 4B single-stream DiT, \method achieves 0.8293 word accuracy on CVTG-2K and 0.938 accuracy on LongText-Bench, closely matching the substantially larger Qwen-Image on both benchmarks. These results demonstrate the value of jointly learned planning representations and region-specific rendering for autonomous visual text generation.
\end{abstract}

\section{Introduction}
\label{sec:intro}

Text is ubiquitous in natural scenes and designed media, from signs and posters to book covers and digital interfaces. Generating text-rich images requires a model to determine what text belongs in the scene, organize multiple regions in space, and faithfully render each string at its intended location~\citep{lin2024parrot,peng2025bizgen,wu2025qwen}. Visual text generation therefore couples content-and-layout planning with rendering in a continuous visual space.

Layout-conditioned diffusion models can synthesize high-quality text-rich images from explicit boxes and strings~\citep{du2025textcrafter,chen2024region}. However, these layouts are typically supplied externally rather than inferred from the prompt. Autonomous generation must instead infer its own layout and use it effectively to guide synthesis.

\begin{figure}[t]
    \centering
    \setlength{\abovecaptionskip}{4pt}
    \setlength{\belowcaptionskip}{2pt}
    \includegraphics[width=0.92\linewidth]{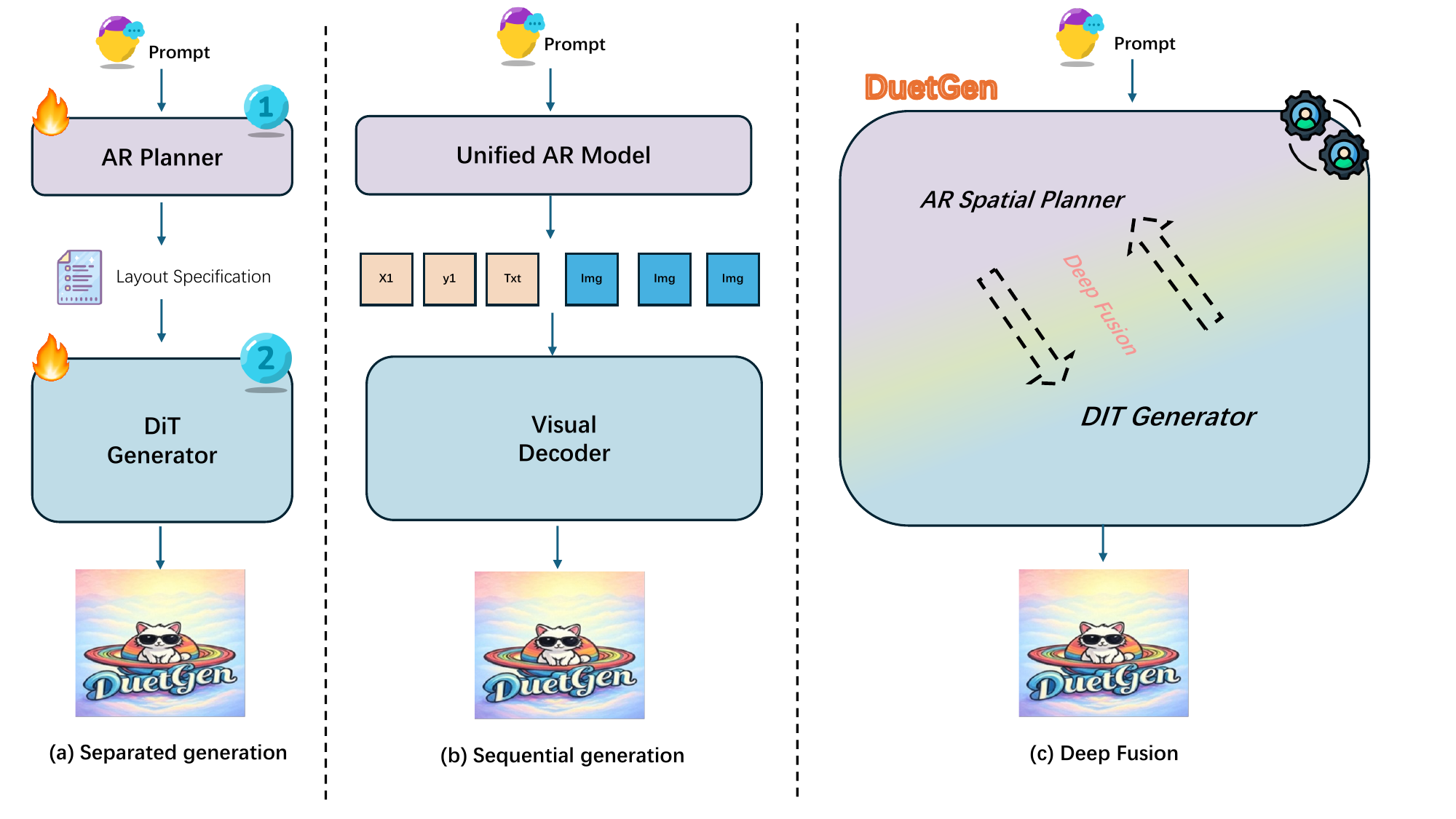}
    \caption{\textbf{Paradigms for autonomous layout-aware visual text generation.}
    (a) Separated AR--diffusion systems separately optimize planning and continuous rendering in two stages.
    (b) Fully autoregressive systems jointly model layout and image tokens, but rely on discrete visual representations.
    (c) DeepFusion jointly optimizes AR planning and continuous diffusion rendering through shared planner states and region-specific execution.}
    \label{fig:paradigm}
    \vspace{-0.35em}
\end{figure}

Two paradigms have been explored toward this goal, as summarized in Figure~\ref{fig:paradigm}. Separated AR--diffusion systems pass an autoregressively generated text--layout specification to a separately optimized diffusion renderer~\citep{chen2024textdiffuser,zhang2025creatilayout}. This two-stage design retains continuous diffusion rendering but lacks deep integration between planning and rendering. Fully autoregressive systems provide tighter integration by modeling layouts and images in a shared sequence~\citep{he2025plangen,lu2025uni,mao2026textground4m}. However, their discrete visual tokens create a representation bottleneck for fine-grained glyph and image synthesis~\citep{yu2023language,ma2025unitok}.

These limitations motivate jointly optimizing structured AR planning and continuous diffusion rendering. The planner representations need to preserve structured content and geometry while serving as effective conditions for synthesis; the DiT must then faithfully execute their region-specific information.

We introduce \method, an autonomous visual text generator that realizes this principle through \textbf{DeepFusion}. An AR planner produces a bbox--content plan, and its prompt and plan hidden states condition a DiT through a lightweight mapper. A joint planning--rendering objective combines autoregressive plan supervision, text-region-weighted diffusion learning, and auxiliary coordinate supervision. Together, these objectives preserve structured planning, improve the spatial precision of plan representations, and focus rendering learning on text-bearing regions. During sampling, Phase-Aware Attention Modulation (PAM) selectively strengthens attention from patches inside each planned region to its matched bbox and content states. DeepFusion thereby integrates planning and continuous rendering through jointly learned representations, coordinated objectives, and region-specific execution.

Our contributions are summarized as follows:
\begin{itemize}
    \setlength{\topsep}{0.1em}
    \setlength{\partopsep}{0pt}
    \setlength{\itemsep}{0.1em}
    \setlength{\parsep}{0pt}
    \item We introduce \method, which deeply integrates autoregressive layout planning and continuous diffusion rendering through DeepFusion.
    \item We jointly learn prompt-and-plan representations through autoregressive plan supervision, text-region-weighted diffusion learning, and auxiliary coordinate supervision, and strengthen region-specific plan execution with PAM.
    \item Extensive analyses and experiments validate this integration across planner scales and diffusion backbones, achieving competitive multi-region and long-text generation.
\end{itemize}

\section{Related Work}
\label{sec:related-work}
\subsection{Visual Text Rendering}
Visual text rendering, which aims to seamlessly integrate legible and controllable text into generated images, has attracted increasing attention. 
Foundational work focused on explicit character-level conditioning, either by incorporating rendered glyphs as priors~\citep{ji2023improving, zhao2024udifftext, ma2023glyphdraw, ma2025glyphdraw2, yang2023glyphcontrol} or by developing specialized character-aware encoders~\mbox{\citep{liu2024glyph, liu2024glyphv2}}. To support multilingual rendering, approaches have leveraged OCR models for stroke features~\citep{tuo2023anytext, tuo2024anytext2} or employed visual glyph replication~\citep{wang2025reptext}. Concurrently, methods emerged to control spatial arrangement through attention map manipulation~\citep{xie2023boxdiff} and full layout-to-image synthesis~\citep{zheng2023layoutdiffusion}.
More recent advances have pursued higher typographic fidelity and novel generation paradigms. These include efforts to enhance word-level fidelity, style, and alignment~\citep{wang2025dreamtext, shi2025fonts, shi2025wordcon}, as well as the exploration of layout-agnostic~\citep{zhangli2024layout} and training-free, outline-guided generation~\citep{zhang2024brush}. 
The problem's scope has also expanded to multi-instance scenarios using agent-based systems~\citep{li2025mccd, zhao2024diffagent} and fine-grained attribute control~\citep{zhou2025dreamrenderer, zhou2024migc}. 
This has led to a surge in domain-specific applications, including specialized models for multilingual text~\citep{jiang2025controltext, xie2025textflux, li2024joytype}, posters~\citep{hu2025dreamposter, gao2025postermaker, chen2025postercraft}, long-form content~\citep{wang2025beyond}, ancient scripts~\citep{li2025oraclefusion}, and broader graphic design automation~\citep{liang2024textcengen, wang2025designdiffusion, chen2025rethinking}.

These prior works primarily focus on the synthesis stage, assuming layout specifications are provided externally. \method, in contrast, operates end-to-end: it autonomously plans a structured blueprint from a prompt and then renders it, using joint training to better align planning with synthesis.

\subsection{Unified Multimodal Modeling}
Unified multimodal modeling has evolved towards single architectures handling both understanding and generation by combining autoregressive models for text with diffusion models for images. Transfusion~\citep{zhou2024transfusion} pioneered this by training a single Transformer on discrete text tokens with causal attention and continuous image patches with bidirectional attention, applying next-token prediction and diffusion objectives simultaneously. Show-o~\citep{xie2024show} builds upon pre-trained LLMs and employs discrete denoising diffusion for image tokens, processing text autoregressively and images via full attention. 
BAGEL~\citep{deng2025emerging} employs a decoder-only architecture with dual Transformer experts for understanding and generation, using separate encoders for semantic content and pixel-level information, while BLIP3-o~\citep{chen2025blip3} uses a diffusion transformer to generate CLIP image features with sequential pretraining. TransDiff~\citep{zhen2025marrying} adopts joint end-to-end training with an AR Transformer encoding inputs into semantic features that guide a diffusion decoder. 

Unlike these general-purpose models, \method is the first to architecturally specialize the AR-Diffusion paradigm for the distinct challenges of text rendering, employing a \textbf{deep fusion} of layout planning with generative synthesis to ensure high-fidelity text rendering.
\section{Method}
\label{sec:method}

\begin{figure*}[t]
    \centering
    \setlength{\abovecaptionskip}{4pt}
    \setlength{\belowcaptionskip}{2pt}
    \includegraphics[width=\textwidth]{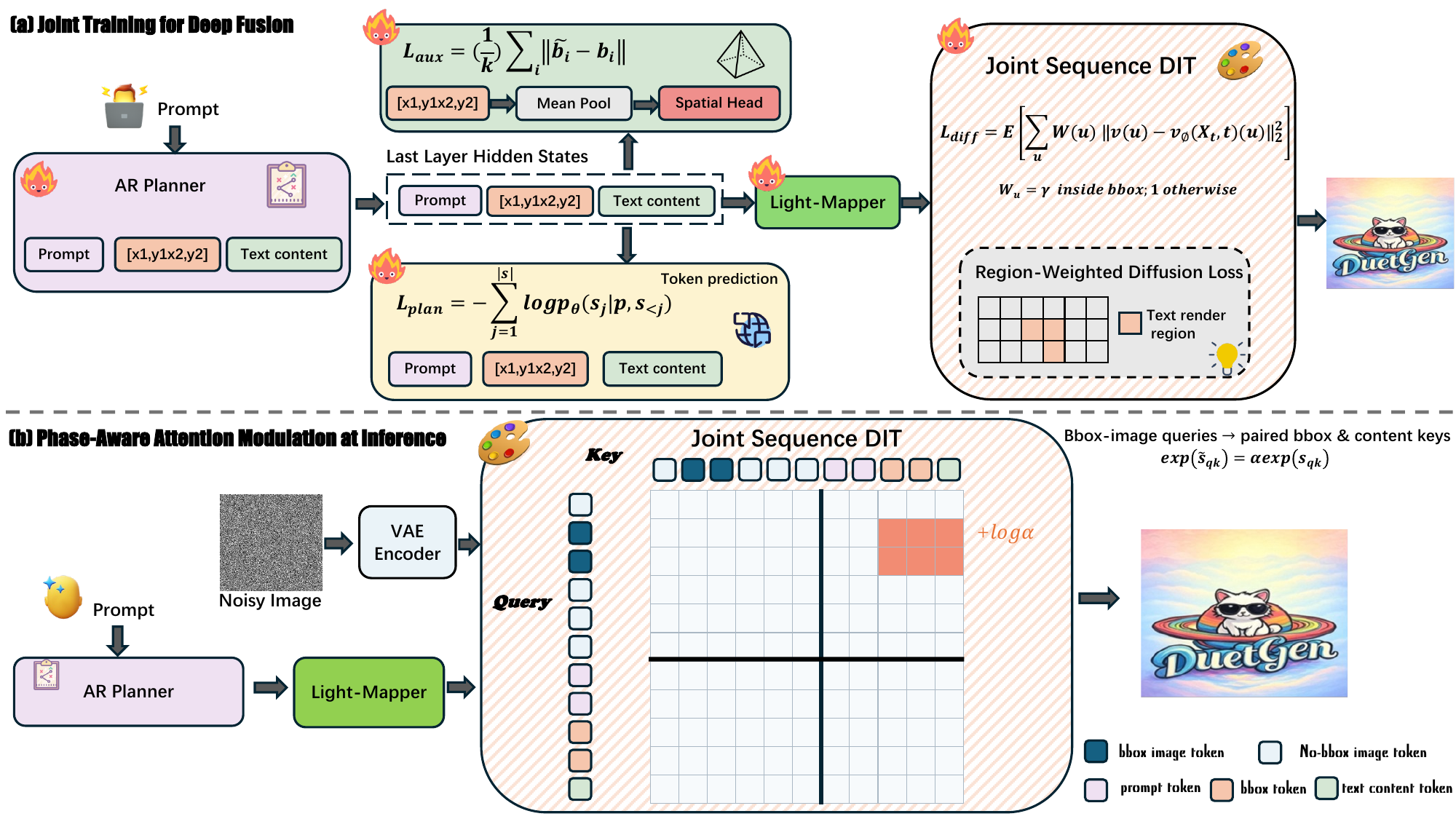}
    \caption{\textbf{Overview of \method.}
    (a) DeepFusion jointly optimizes planning and rendering through prompt-and-plan states, autoregressive plan supervision, text-region-weighted diffusion learning, and auxiliary coordinate supervision.
    (b) During inference, Phase-Aware Attention Modulation strengthens attention from in-region patches to their matched bbox and content states in middle DiT layers and denoising steps.}
    \label{fig:framework}
    \vspace{-0.35em}
\end{figure*}

\method couples a Qwen3.5-2B autoregressive planner~\citep{qwen3.5} with a 4B single-stream DiT through DeepFusion. We pretrain the DiT from random initialization for general image generation before joint planning--rendering training (Appendix~\ref{sec:appendix-dit-pretraining}). Given a prompt $p$, the planner emits a bbox--content plan specifying the position and content of each text region. A lightweight mapper projects the planner's final-layer prompt and plan states into condition tokens for the DiT. The DiT then denoises image latents while attending jointly to these condition tokens. Figure~\ref{fig:framework} summarizes training and inference.

\subsection{Bbox--Content Planning}
\label{sec:layout-reasoning}

The plan is $\layout=\{(b_i,c_i)\}_{i=1}^{K}$, where
$b_i=(x_i^1,y_i^1,x_i^2,y_i^2)$ is a bounding box and $c_i$ is the text to be rendered inside it. We serialize each region as
\begin{equation}
    s_i=\big[
    \langle\mathrm{region}\rangle,\,
    x_i^1,y_i^1,x_i^2,y_i^2,\,
    \langle\mathrm{text}\rangle,\,
    c_i,\,
    \langle/\mathrm{region}\rangle
    \big].
    \label{eq:region-serialization}
\end{equation}
We refer to $s_i$ as a \emph{bbox--content sequence}: the four coordinate fields form the coordinate span, and the tokens following $\langle\mathrm{text}\rangle$ form the content span.

Let $s=(s_1,\ldots,s_K)$ denote the complete serialized plan. During training, the planner is teacher-forced on the ground-truth plan; at inference, it generates the plan autoregressively. Its LM head maps each hidden state to the next-token distribution $p_\theta$, yielding the autoregressive plan loss
\begin{equation}
    \mathcal{L}_{\mathrm{plan}}
    =-\sum_{j=1}^{|s|}
    \log p_\theta(s_j\mid p,s_{<j}).
    \label{eq:plan-loss}
\end{equation}

\subsection{Joint Training}
\label{sec:joint-training}

We concatenate the prompt and bbox--content sequence and extract the planner's final-layer hidden states $\mathbf H_p$ and $\mathbf H_s$ at their respective token positions. A lightweight mapper $m_\eta$ converts these prompt-and-plan states into DiT condition tokens $\mathbf C$:
\begin{equation}
    \mathbf C=m_\eta([\mathbf H_p;\mathbf H_s]),
    \qquad
    \mathbf X_t=[\mathbf Z_t;\mathbf C].
    \label{eq:joint-sequence}
\end{equation}
$\mathbf Z_t$ denotes the patch tokens of $\mathbf z_t$. The DiT processes $\mathbf X_t$ as a single sequence.

For flow-matching training, we sample
$\mathbf z_t=(1-\sigma_t)\mathbf z_0+\sigma_t\epsilon$ and predict the velocity
$v=\epsilon-\mathbf z_0$. We apply a text-region-weighted diffusion loss:
\begin{equation}
    \mathcal{L}_{\mathrm{diff}}
    =\mathbb E\!\left[
    \sum_{\mathbf u}W(\mathbf u)
    \left\|
    v(\mathbf u)-v_\phi(\mathbf X_t,t)(\mathbf u)
    \right\|_2^2
    \right].
    \label{eq:diffusion-loss}
\end{equation}
Here, $\mathbf u$ indexes a spatial location on the latent grid, and $W(\mathbf u)$ equals $\gamma$ on the union of rasterized reference boxes and $1$ elsewhere. The velocity prediction is evaluated on this grid after unpatchifying the DiT output. Setting $\gamma=1$ gives the standard flow-matching loss; Appendix~\ref{sec:appendix-flow-loss} specifies the rasterization.

For auxiliary coordinate supervision, we mean-pool the states of each coordinate span and regress its normalized box:
\begin{equation}
    \tilde b_i=
    \sigma\!\left(g_\psi(\operatorname{Mean}(\mathbf H_{b_i}))\right),
    \qquad
    \mathcal{L}_{\mathrm{aux}}
    =\frac{1}{K}\sum_{i=1}^{K}\|\tilde b_i-b_i^{\mathrm{norm}}\|_1,
    \label{eq:auxiliary-loss}
\end{equation}
where $\mathbf H_{b_i}\subset\mathbf H_s$ denotes the coordinate-span states of region $i$, and $b_i^{\mathrm{norm}}\in[0,1]^4$ is its reference box normalized by image width and height. The auxiliary head $g_\psi$ is used only during training. The prompt-and-plan states, rather than the box predictions from this head, condition the DiT.

The joint planning--rendering objective is
\begin{equation}
    \mathcal L
    =\mathcal L_{\mathrm{plan}}
    +\lambda_{\mathrm{diff}}\mathcal L_{\mathrm{diff}}
    +\lambda_{\mathrm{aux}}\mathcal L_{\mathrm{aux}}.
    \label{eq:full-objective}
\end{equation}
We jointly optimize the planner, mapper, DiT, and auxiliary head, while keeping the VAE frozen.

\subsection{Phase-Aware Attention Modulation}
\label{sec:region-binding}

At inference, Phase-Aware Attention Modulation (PAM) strengthens only the attention from image patches inside each planned region to the bbox and text-content tokens of the same region. The modulation is restricted to the middle DiT layers $\mathcal M$ and a denoising window $\mathcal T_{\mathrm{mid}}$:
\begin{equation}
\begin{aligned}
    \widetilde s_{qk}^{(\ell,t)}
    &=\frac{\langle Q_q^{(\ell,t)},K_k^{(\ell,t)}\rangle}{\sqrt d}
    +\Delta_{qk}^{(\ell,t)},\\
    \Delta_{qk}^{(\ell,t)}&=
    \begin{cases}
        \log\alpha,
        & \ell\in\mathcal M,\ t\in\mathcal T_{\mathrm{mid}},
          \ q\in\mathcal P_i^{\mathrm{box}},
          k\in\mathcal K_i^{\mathrm{box}}\cup
          \mathcal K_i^{\mathrm{text}}
          \text{ for some } i,\\
        0, & \text{otherwise}.
    \end{cases}
\end{aligned}
    \label{eq:binding-attention}
\end{equation}
$\mathcal P_i^{\mathrm{box}}$ denotes the indices of DiT image tokens whose patch centers lie inside $b_i$; for each $q\in\mathcal P_i^{\mathrm{box}}$, $Q_q^{(\ell,t)}$ is the corresponding query vector. $\mathcal K_i^{\mathrm{box}}$ and $\mathcal K_i^{\mathrm{text}}$ index the condition-token positions projected from the coordinate and content spans of region $i$. For $\alpha>1$, adding $\log\alpha$ multiplies the unnormalized attention weights of exactly these query--key pairs by $\alpha$. All other logits are unchanged. Softmax renormalizes each affected query row, so normalized attention to other keys in that row can decrease. PAM thus increases the relative attention assigned to region-matched bbox and content tokens.

\section{Experiments}
\label{sec:experiments}

\subsection{Experimental Setup}
\label{sec:experimental-setup}

\paragraph{Datasets.}
We first pretrain the DiT on 600K generic captioned images at $512\times512$ resolution. For joint training, we combine MARIO-10M~\citep{chen2023textdiffuser}, CreateDesign~\citep{zhang2025creatidesign}, Lex-10K~\citep{zhao2025lex}, PrismLayers~\citep{chen2025prismlayers}, and LLaVAR-2~\citep{zhou2024high}; filtering and OCR-based processing yield 6,266,861 text--image pairs with transcriptions and bounding boxes. We evaluate on CVTG-2K~\citep{du2025textcrafter}, LongText-Bench~\citep{geng2025x}, and TextAtlasEval~\citep{wang2025textatlas5m}.

\paragraph{Implementation details.}
We initialize the planner with Qwen3.5-2B~\citep{qwen3.5} and use a 6B joint-sequence DiT. The complete model is jointly trained for one epoch on 16 NVIDIA A800 GPUs using bfloat16 precision, gradient checkpointing, and DeepSpeed ZeRO-2~\citep{rajbhandari2020zero}, with a global batch size of 64. We use AdamW~\citep{loshchilov2017decoupled}: the planner uses a learning rate of $1\times10^{-5}$, while the DiT, mapper, and auxiliary head use $5\times10^{-5}$. We set the weight decay to 0.01 and apply cosine decay with a 3\% warmup ratio. Joint training and inference use $1024\times1024$ images. We set $\lambda_{\mathrm{diff}}=1$, $\lambda_{\mathrm{aux}}=0.1$, and the text-region weight in Eq.~\ref{eq:diffusion-loss} to $\gamma=1.2$. Inference uses 40 denoising steps. PAM is applied to the middle half of DiT layers from steps 20 to 30, with attention amplification factor $\alpha=1.2$.


\subsection{Main Results}
\label{sec:main-results}

\paragraph{Quantitative results.}
On CVTG-2K (Table~\ref{tab:cvtg2k}), \method reaches 0.8293 average word accuracy, closely matching the 20B Qwen-Image~\citep{wu2025qwen} (0.8301), and performs best on the most crowded five-region split. On LongText-Bench (Table~\ref{tab:longtext}), it attains 0.938 accuracy, within 0.005 of Qwen-Image~\citep{wu2025qwen} and above all other baselines. These results show that \method approaches the text-rendering accuracy of a substantially larger model with fewer parameters and less training data.

\begin{table}[!t]
    \vspace*{-10pt}
    \caption{Quantitative results on CVTG-2K~\citep{du2025textcrafter}. Word accuracy is grouped by the number of text regions. Best and second-best results are \textbf{bold} and \underline{underlined}.}
    \label{tab:cvtg2k}
    \centering
    \setlength{\tabcolsep}{2.5pt}
    \resizebox{\linewidth}{!}{%
    \begin{tabular}{l|ccccc|cc}
        \toprule
        \multirow{2}{*}{\textcolor{tblInk}{\textbf{Method}}} & \multicolumn{5}{c|}{\cellcolor{tblBlue}\textcolor{tblInk}{\textbf{Word Accuracy $\uparrow$}}} & \multirow{2}{*}{\textcolor{tblGood}{\textbf{NED $\uparrow$}}} & \multirow{2}{*}{\textcolor{tblWarm}{\textbf{CLIPScore $\uparrow$}}} \\
        \cmidrule(lr){2-6}
        & \cellcolor{tblBlue}\textbf{2} & \cellcolor{tblBlue}\textbf{3} & \cellcolor{tblBlue}\textbf{4} & \cellcolor{tblBlue}\textbf{5} & \cellcolor{tblTeal}\textbf{Avg.} & & \\
        \midrule
        SD3.5 Large~\citep{esser2024scaling} & 0.7293 & 0.6825 & 0.6574 & 0.5940 & 0.6548 & 0.8470 & 0.7797 \\
        \rowcolor{tblStripe}
        FLUX.1 dev~\citep{flux} & 0.6089 & 0.5531 & 0.4661 & 0.4316 & 0.4965 & 0.6879 & 0.7401 \\
        AnyText~\citep{tuo2023anytext} & 0.0513 & 0.1739 & 0.1948 & 0.2249 & 0.1804 & 0.4675 & 0.7432 \\
        \rowcolor{tblStripe}
        TextDiffuser-2~\citep{chen2024textdiffuser} & 0.5322 & 0.3255 & 0.1787 & 0.0809 & 0.2326 & 0.4353 & 0.6765 \\
        RAG-Diffusion~\citep{chen2024region} & 0.4388 & 0.3316 & 0.2116 & 0.1910 & 0.2648 & 0.4498 & 0.7797 \\
        \rowcolor{tblStripe}
        3DIS~\citep{zhou20243dis} & 0.4495 & 0.3959 & 0.3880 & 0.3303 & 0.3813 & 0.6505 & 0.7767 \\
        TextCrafter~\citep{du2025textcrafter} & 0.7628 & 0.7628 & 0.7406 & 0.6977 & 0.7370 & 0.8679 & 0.7868 \\
        \rowcolor{tblStripe}
        Seedream 3.0~\citep{gao2025seedream} & 0.6282 & 0.5962 & 0.6043 & 0.5610 & 0.5924 & 0.8537 & 0.7821 \\
        Show-o2~\citep{xie2025show} & 0.1715 & 0.1358 & 0.1193 & 0.0681 & 0.1236 & 0.4152 & 0.7204 \\
        \rowcolor{tblStripe}
        Janus-Pro~\citep{chen2025janus} & 0.1791 & 0.1412 & 0.1275 & 0.0736 & 0.1303 & 0.4278 & 0.7329 \\
        BLIP3o-Next~\citep{chen2025blip3o} & 0.3821 & 0.3415 & 0.3275 & 0.1736 & 0.3061 & 0.4898 & 0.7319 \\
        \rowcolor{tblTeal}
        Qwen-Image~\citep{wu2025qwen} & \tblsecond{0.8370} & \tblbest{0.8364} & \tblbest{0.8313} & \tblsecond{0.8158} & \tblbest{0.8301} & \tblbest{0.9116} & \tblbest{0.7982} \\
        \rowcolor{tblLilac}
        \textbf{\method} & \tblbest{0.8376} & \tblsecond{0.8321} & \tblsecond{0.8309} & \tblbest{0.8177} & \tblsecond{0.8293} & \tblsecond{0.9113} & \tblsecond{0.7931} \\
        \bottomrule
    \end{tabular}%
    }

    \vspace*{-3pt}
\end{table}

\begin{table}[t]
    \caption{Text accuracy on LongText-Bench~\citep{geng2025x}. Best and second-best results are \textbf{bold} and \underline{underlined}.}
    \label{tab:longtext}
    \centering
    \small
    \setlength{\tabcolsep}{5pt}
    \begin{tabular}{lc|lc}
        \toprule
        \rowcolor{tblBlue}
        \textcolor{tblInk}{\textbf{Method}} & \textcolor{tblInk}{\textbf{Acc. $\uparrow$}} & \cellcolor{tblTeal}\textcolor{tblGood}{\textbf{Method}} & \cellcolor{tblTeal}\textcolor{tblGood}{\textbf{Acc. $\uparrow$}} \\
        \midrule
        BLIP3-o~\citep{chen2025blip3} & 0.021 & OmniGen2~\citep{wu2025omnigen2} & 0.561 \\
        \rowcolor{tblStripe}
        Show-o2~\citep{xie2025show} & 0.119 & FLUX.1~\citep{flux} & 0.607 \\
        Janus-Pro~\citep{chen2025janus} & 0.019 & TextDiffuser-2~\citep{chen2024textdiffuser} & 0.371 \\
        \rowcolor{tblStripe}
        Kolors 2.0~\citep{Kolors2} & 0.543 & HiDream-I1-Full~\citep{cai2025hidream} & 0.543 \\
        RAG-Diffusion~\citep{chen2024region} & 0.321 & TextCrafter~\citep{du2025textcrafter} & 0.683 \\
        \rowcolor{tblTeal}
        Qwen-Image~\citep{wu2025qwen} & \tblbest{0.943} & \cellcolor{tblLilac}\textbf{\method} & \cellcolor{tblLilac}\tblsecond{0.938} \\
        \bottomrule
    \end{tabular}
    \vspace*{-4pt}
    
\end{table}

\begin{figure}[t]
    \centering
    \includegraphics[width=\linewidth]{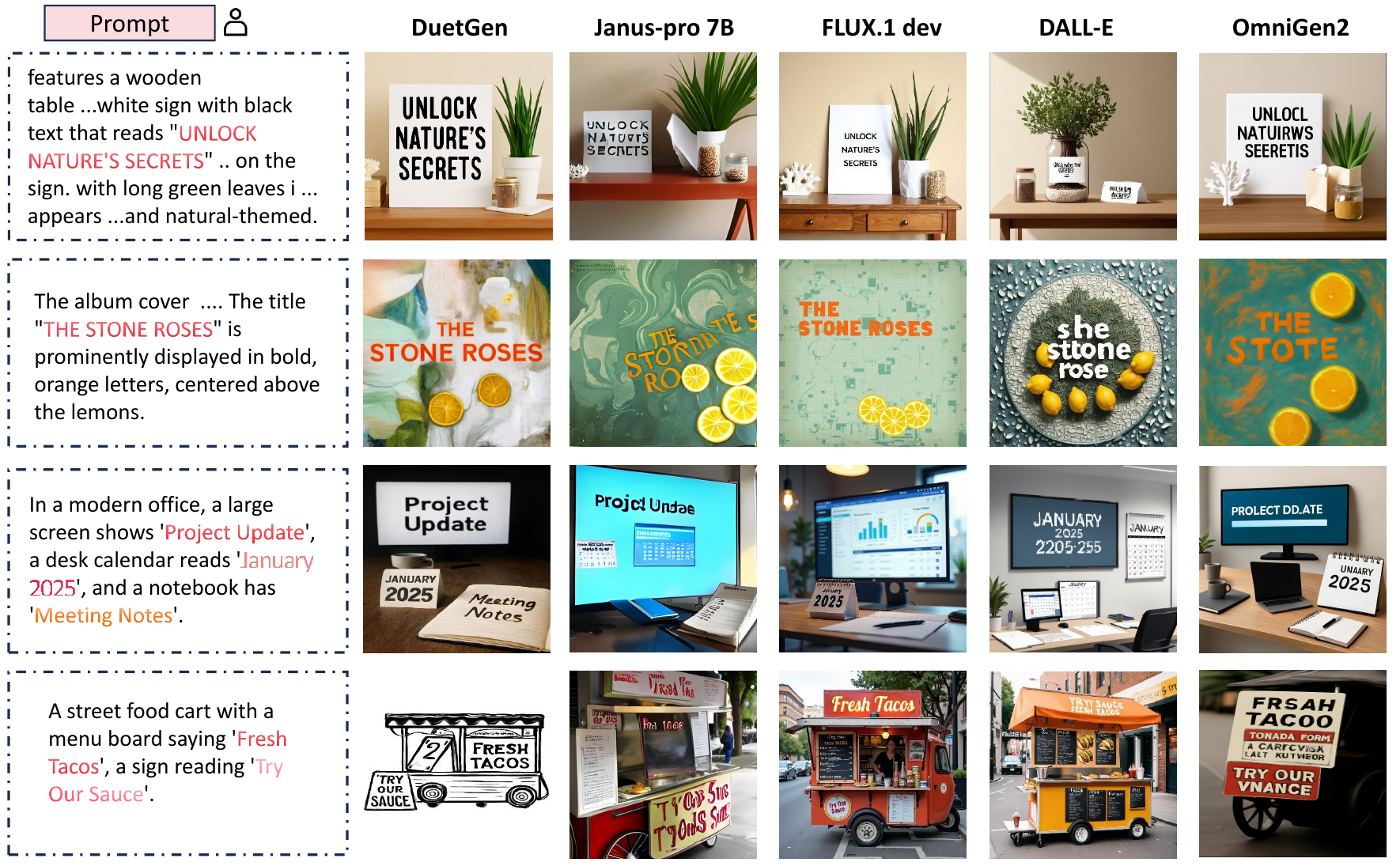}
    \caption{Qualitative comparison on prompts containing multiple text regions and extended text.}
    \label{fig:qualitative-main}
\end{figure}

\paragraph{Qualitative results.}
Figure~\ref{fig:qualitative-main} compares representative multi-region and long-text generations with Janus-Pro~\citep{chen2025janus}, FLUX.1~\citep{flux}, DALL-E 3~\citep{betker2023improving}, and OmniGen2~\citep{wu2025omnigen2}. The examples jointly test text content, regional placement, and visual composition. Across these cases, \method preserves the requested phrases and their intended spatial organization more consistently, while the comparison models exhibit missing or corrupted characters and region-placement errors.

\section{DeepFusion Analysis}
\label{sec:deepfusion-analysis}

We analyze DeepFusion along the planning-to-rendering path: whether coordinate-span states encode box geometry, how the joint planning--rendering objective shapes this information, and how the diffusion-side mechanisms execute region-specific conditions. We further compare joint and separate planner training and assess the transfer of the diffusion-side mechanisms across DiT backbones. Unless otherwise stated, all variants follow the settings in Section~\ref{sec:experimental-setup}.

\subsection{Spatial Information at the AR--DiT Interface}
\label{sec:interface-analysis}

\subsubsection{Do Coordinate-Span Representations Encode Box Geometry?}
\label{sec:native-spatial-representation}

We probe the planner's final-layer states associated with each coordinate span before their projection into the DiT condition tokens in Eq.~\ref{eq:joint-sequence}. We mean-pool the states within each span and fit a post-hoc ridge regressor to predict $(x_1,y_1,x_2,y_2)$. Evaluation uses held-out prompts containing 1,034 boxes and reports mean $R^2$, pixel RMSE, and median pixel error at $1024\times1024$ resolution.

\begin{figure*}[t]
    \centering
    \begin{minipage}[t]{0.49\textwidth}
        \centering
        \includegraphics[width=\linewidth]{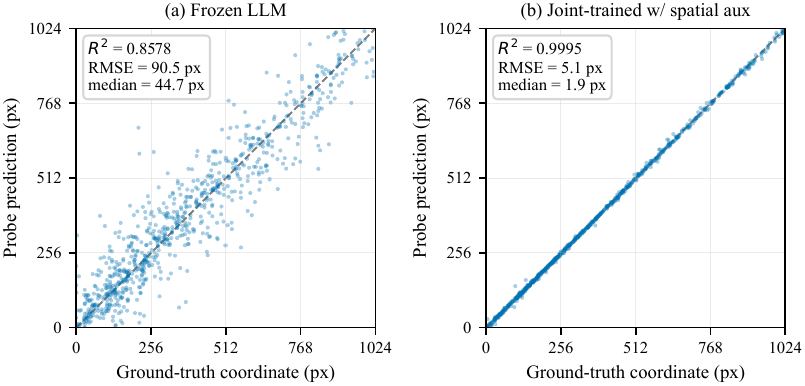}
        \captionof{figure}{Linear readout of coordinate-span states in the pretrained planner (left) and after optimization with the full objective (right).}
        \label{fig:spatial-probe-scatter}
    \end{minipage}\hfill
    \begin{minipage}[t]{0.49\textwidth}
        \centering
        \includegraphics[width=\linewidth]{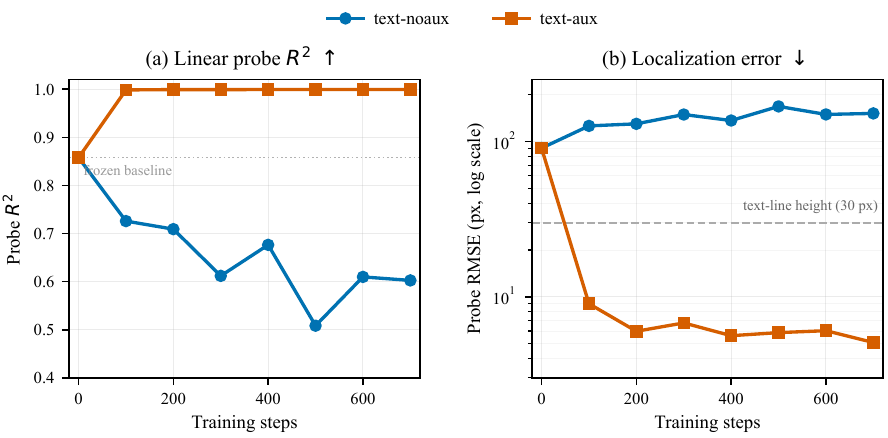}
        \captionof{figure}{Auxiliary supervision restores spatial readability after diffusion-only degradation.}
        \label{fig:spatial-probe-curve}
    \end{minipage}
    \par\vspace{-0.5em}
\end{figure*}

The frozen pretrained planner already encodes linearly decodable box geometry, achieving $R^2=0.8578$, 90.5-pixel RMSE, and 44.7-pixel median error in Figure~\ref{fig:spatial-probe-scatter}(a). Updating the planner with the full joint planning--rendering objective improves these results to $R^2=0.9995$, 5.1-pixel RMSE, and 1.9-pixel median error (Figure~\ref{fig:spatial-probe-scatter}(b)). Thus, the coordinate-span states used at the AR--DiT interface encode box geometry, whose linear readability is substantially improved under the full objective.

\subsubsection{Is the Diffusion Objective Sufficient?}
\label{sec:diffusion-gradient-analysis}

We isolate the effects of different objectives on planner optimization while keeping DiT training fixed. The planner is either frozen or updated by $\mathcal{L}_{\mathrm{diff}}$, $\mathcal{L}_{\mathrm{diff}}+\mathcal{L}_{\mathrm{plan}}$, or the full joint objective. Each checkpoint is evaluated with an independently fitted linear probe.

Diffusion-only training degrades spatial readability, reducing $R^2$ from 0.8578 to 0.6026 and increasing RMSE from 90.5 to 151.5 pixels. Adding autoregressive plan supervision provides only marginal recovery ($R^2=0.6129$, RMSE $=147.3$). Adding auxiliary coordinate supervision instead improves the probe to $R^2=0.9995$ and 5.1-pixel RMSE.

The same pattern appears in planning--rendering consistency (PRC), measured as the mean IoU between planned and rendered text boxes. On the four-region CVTG-2K subset~\citep{du2025textcrafter}, PRC decreases from 0.4892 with a frozen planner to 0.3121 under diffusion-only training, while adding the plan loss recovers it only to 0.3469 (Table~\ref{tab:aux-generation-bridge}). The full objective raises PRC to 0.9176. These results connect auxiliary coordinate supervision to both the spatial readability of coordinate-span states and the accuracy with which the DiT executes the generated layout.

\subsubsection{Does Spatial Readability Depend on Pretrained Digit Tokens?}
\label{sec:coordinate-encoding}

Having established the role of auxiliary coordinate supervision, we test whether its effect depends on the pretrained numerical structure of digit tokens. We compare the digit serialization used by \method, such as $[235,45,787,71]$, with a newly initialized coordinate-bin vocabulary containing 64 tokens per axis. Each bin covers 16 pixels at $1024\times1024$ resolution; for example, the same box is represented as $\langle\mathrm{xbin}{14}\rangle\langle\mathrm{ybin}{2}\rangle\langle\mathrm{xbin}{49}\rangle\langle\mathrm{ybin}{4}\rangle$.

Under the same coordinate-supervised setup, the randomly initialized bin tokens become linearly readable after approximately 150--200 steps and converge to a 6.4-pixel RMSE, close to the 5.1 pixels obtained with pretrained digits (Figure~\ref{fig:coordinate-encoding-probe}). Thus, linearly readable spatial representations do not require pretrained digit tokens; digit serialization mainly accelerates convergence and yields slightly lower final error.

\subsubsection{Can Planner Scale Replace Joint Optimization?}
\label{sec:joint-versus-scale}

We test whether planner scale can replace joint optimization. We compare the jointly optimized Qwen3.5-2B planner with separately fine-tuned Qwen3.5-2B, Llama3.1-8B~\citep{llama3.1}, and Qwen3.5-9B~\citep{qwen3.5} planners under matched layout supervision, renderer initialization, and data. Sep.\ SFT freezes the planner and trains only the mapper--DiT with $\mathcal L_{\mathrm{diff}}$; joint training also updates the planner with it. On the four-region CVTG-2K subset, we report word accuracy, PRC, and overlap rate (OR).

The same-size separate control reaches only 0.6341 accuracy, 0.6592 PRC, and 0.116 OR (Table~\ref{tab:joint-versus-scale}); even the separate 9B planner remains below the jointly optimized 2B planner. Thus, neither separate training nor planner scaling recovers the planning--rendering alignment achieved by joint optimization.

\begin{table*}[t]
    \centering

    \begin{minipage}[t]{0.56\textwidth}
        \centering
        \captionof{table}{
            Planner-update effects on spatial readability and generation
            on the four-region CVTG-2K subset.
        }
        \label{tab:aux-generation-bridge}

        \scriptsize
        \setlength{\tabcolsep}{1.8pt}
        \resizebox{\linewidth}{!}{%
        \begin{tabular}{l|ccc|ccc}
            \toprule

            \multirow{2}{*}{\textcolor{tblInk}{\textbf{Planner update}}}
            & \multicolumn{3}{c|}{\cellcolor{tblBlue}\textcolor{tblInk}{\textbf{Spatial probe}}}
            & \multicolumn{3}{c}{\cellcolor{tblTeal}\textcolor{tblGood}{\textbf{Generation}}} \\
            \cmidrule(lr){2-4}
            \cmidrule(lr){5-7}

            & \cellcolor{tblBlue}$R^2\uparrow$
            & \cellcolor{tblBlue}RMSE$_{\mathrm{px}}\downarrow$
            & \cellcolor{tblBlue}Med.$_{\mathrm{px}}\downarrow$
            & \cellcolor{tblTeal}Acc.$\uparrow$
            & \cellcolor{tblTeal}PRC$\uparrow$
            & \cellcolor{tblTeal}CLIP$\uparrow$ \\

            \midrule

            \rowcolor{tblStripe}
            Frozen
            & 0.8578
            & 90.5
            & 44.7
            & 0.7032
            & 0.4892
            & 0.6987 \\

            \rowcolor{tblPeach}
            $\mathcal L_{\mathrm{diff}}$
            & 0.6026
            & 151.5
            & 66.4
            & 0.6788
            & 0.3121
            & 0.7011 \\

            $\mathcal L_{\mathrm{diff}}+\mathcal L_{\mathrm{plan}}$
            & 0.6129
            & 147.3
            & 65.7
            & 0.6832
            & 0.3469
            & 0.7187 \\

            \rowcolor{tblLilac}
            $\mathcal L_{\mathrm{diff}}+\mathcal L_{\mathrm{plan}}+\mathcal L_{\mathrm{aux}}$
            & \tblbest{0.9995}
            & \tblbest{5.1}
            & \tblbest{1.9}
            & \tblbest{0.8309}
            & \tblbest{0.9176}
            & \tblbest{0.7925} \\

            \bottomrule
        \end{tabular}%
        }
    \end{minipage}
    \hfill
    \begin{minipage}[t]{0.42\textwidth}
        \centering
        \captionof{table}{
            Jointly optimized 2B planner versus same-size and larger separately trained planners. Lower OR indicates fewer conflicts among planned boxes.
        }
        \label{tab:joint-versus-scale}

        \scriptsize
        \setlength{\tabcolsep}{2.4pt}
        \renewcommand{\arraystretch}{1.20}
        \resizebox{\linewidth}{!}{%
        \begin{tabular}{l|cc|ccc}
            \toprule

            \rowcolor{tblBlue}
            \textcolor{tblInk}{\textbf{Planner}}
            & \cellcolor{tblPeach}\textbf{Train.}
            & \cellcolor{tblPeach}\textbf{Params}
            & \cellcolor{tblTeal}Acc.$\uparrow$
            & \cellcolor{tblTeal}PRC$\uparrow$
            & \cellcolor{tblTeal}OR$\downarrow$ \\

            \midrule

            \rowcolor{tblLilac}
            \method planner
            & \textcolor{tblGood}{Joint}
            & 2B
            & \tblbest{0.8309}
            & \tblbest{0.9176}
            & \tblbest{0.063} \\

            \rowcolor{tblStripe}
            Qwen3.5-2B
            & Sep.\ SFT
            & 2B
            & 0.6341
            & 0.6592
            & 0.116 \\

            Llama3.1-8B
            & Sep.\ SFT
            & 8B
            & 0.6783
            & 0.6624
            & 0.129 \\

            \rowcolor{tblStripe}
            Qwen3.5-9B
            & Sep.\ SFT
            & 9B
            & 0.6936
            & 0.7189
            & 0.114 \\

            \bottomrule
        \end{tabular}%
        }
    \end{minipage}
    \par\vspace{-0.4em}
\end{table*}

\subsection{Region-Specific Execution}
\label{sec:component-ablation}

Building on the spatial analysis above, we evaluate two diffusion-side mechanisms for region-specific execution: text-region-weighted diffusion training and PAM at inference.

\paragraph{Text-region-weighted diffusion learning.}
Table~\ref{tab:diffusion-side-analysis}(a) varies the text-region weight $\gamma$ on the StyledTextSynth evaluation subset of TextAtlasEval~\citep{wang2025textatlas5m}, with PAM enabled. Increasing $\gamma$ from 1.0 to 1.2 reduces FID from 74.15 to 68.70 and raises OCR accuracy from 48.32 to 69.93. Larger weights reverse part of these gains, with $\gamma=1.2$ achieving the best result across all three metrics.

\begin{figure*}[t]
    \centering
    \begin{minipage}[t]{0.49\textwidth}
        \centering
        \includegraphics[width=\linewidth]{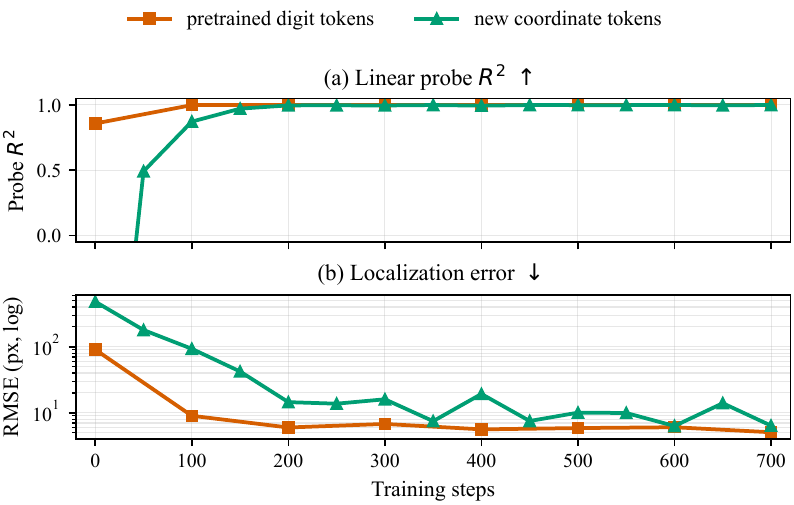}
        \captionof{figure}{Spatial readability of pretrained digit and newly initialized coordinate-bin tokens.}
        \label{fig:coordinate-encoding-probe}
    \end{minipage}\hfill
    \begin{minipage}[t]{0.49\textwidth}
        \centering
        \includegraphics[width=\linewidth]{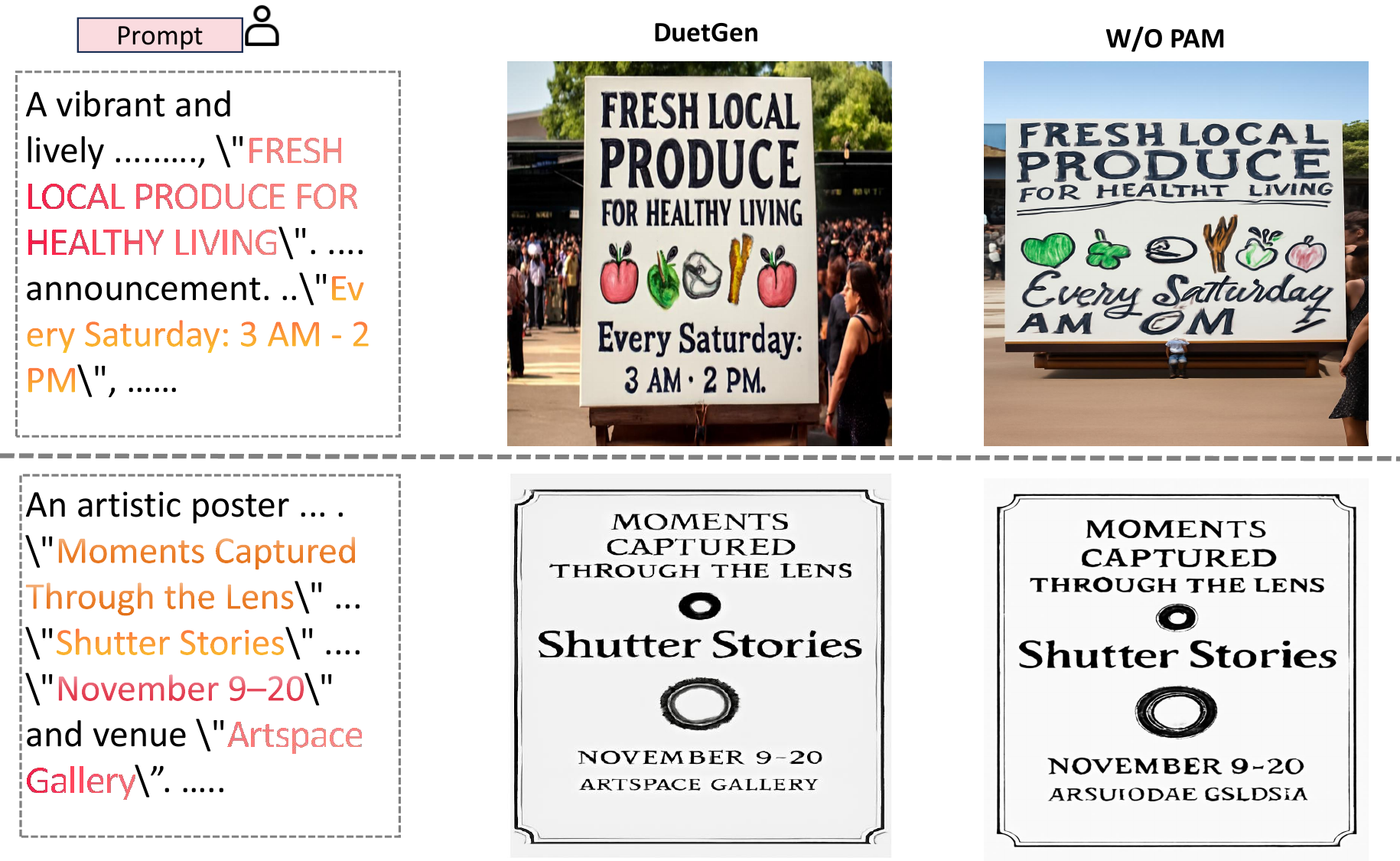}
        \captionof{figure}{Qualitative comparison with and without PAM.}
        \label{fig:region-binding-effect}
    \end{minipage}
    \par\vspace{0.25em}
\end{figure*}

\paragraph{Phase-Aware Attention Modulation.}
PAM selectively strengthens attention from image tokens inside each planned box to the region-matched bbox and content tokens, restricting the modulation to the middle DiT layers and denoising phase. Figure~\ref{fig:region-binding-effect} qualitatively shows that PAM more faithfully renders the requested strings across multiple text regions.

On the same evaluation subset, with $\gamma=1.2$, PAM increases OCR accuracy from 61.71 to 69.93, reduces FID from 72.03 to 68.70, and improves CLIPScore from 0.3168 to 0.3271 (Table~\ref{tab:diffusion-side-analysis}(a)). Table~\ref{tab:diffusion-side-analysis}(c) further localizes these gains: the middle denoising phase and middle layer range perform best in their respective sweeps. Applying PAM to all steps or layers retains part of the improvement but underperforms the localized setting. Increasing $\alpha$ beyond 1.2 similarly reverses part of the gain. These results support PAM as a localized attention modulation rather than uniform amplification throughout denoising.

\begin{table*}[t]
    \centering
    \setlength{\abovecaptionskip}{2pt}
    \setlength{\belowcaptionskip}{2pt}
    \renewcommand{\arraystretch}{0.92}
    \caption{
        Diffusion-side ablations on the StyledTextSynth evaluation subset of TextAtlasEval:
        (a) text-region weighting and PAM, (b) backbone transfer, and (c) PAM design.
        CS and Acc.\ denote CLIPScore and OCR accuracy.
    }
    \label{tab:diffusion-side-analysis}

    \begin{minipage}[t]{0.39\textwidth}
        \centering
        \textbf{(a) Main DiT ablations}\\[2pt]
        \scriptsize
        \setlength{\tabcolsep}{2.0pt}
        \resizebox{\linewidth}{!}{%
        \begin{tabular}{cc|ccc}
            \toprule
            \rowcolor{tblBlue}
            \textcolor{tblInk}{Study} & \textcolor{tblInk}{Setting}
            & \cellcolor{tblTeal}FID$\downarrow$
            & \cellcolor{tblTeal}CS$\uparrow$
            & \cellcolor{tblTeal}Acc.$\uparrow$ \\
            \midrule

            \multirow{4}{*}{\textcolor{tblSecond}{Region wt.}}
            & $\gamma=1.0$
            & 74.15 & 0.2640 & 48.32 \\

            & \cellcolor{tblLilac}$\gamma=\mathbf{1.2}$
            & \cellcolor{tblLilac}\tblbest{68.70}
            & \cellcolor{tblLilac}\tblbest{0.3271}
            & \cellcolor{tblLilac}\tblbest{69.93} \\

            & $\gamma=1.5$
            & 73.50 & 0.2880 & 54.10 \\

            & $\gamma=2.0$
            & 77.80 & 0.2750 & 55.24 \\

            \midrule

            \multirow{2}{*}{\textcolor{tblWarm}{PAM}}
            & PAM off
            & 72.03 & 0.3168 & 61.71 \\

            & \cellcolor{tblLilac}\textcolor{tblGood}{PAM on}
            & \cellcolor{tblLilac}\tblbest{68.70}
            & \cellcolor{tblLilac}\tblbest{0.3271}
            & \cellcolor{tblLilac}\textbf{69.93} \\

            \bottomrule
        \end{tabular}%
        }
    \end{minipage}
    \hfill
    \begin{minipage}[t]{0.59\textwidth}
        \centering
        \textbf{(b) Transfer across diffusion backbones}\\[2pt]
        \scriptsize
        \setlength{\tabcolsep}{2.2pt}
        \resizebox{\linewidth}{!}{%
        \begin{tabular}{cc|ccc|ccc}
            \toprule
            \multirow{2}{*}{\textcolor{tblInk}{\textbf{Region wt.}}}
            & \multirow{2}{*}{\textcolor{tblInk}{\textbf{PAM}}}
            & \multicolumn{3}{c|}{\cellcolor{tblBlue}\textcolor{tblInk}{\textbf{PixArt-$\Sigma$}}}
            & \multicolumn{3}{c}{\cellcolor{tblTeal}\textcolor{tblGood}{\textbf{SD3}}} \\
            \cmidrule(lr){3-5}
            \cmidrule(lr){6-8}

            & & \cellcolor{tblBlue}FID$\downarrow$
            & \cellcolor{tblBlue}CS$\uparrow$
            & \cellcolor{tblBlue}Acc.$\uparrow$
            & \cellcolor{tblTeal}FID$\downarrow$
            & \cellcolor{tblTeal}CS$\uparrow$
            & \cellcolor{tblTeal}Acc.$\uparrow$ \\

            \midrule

            \rowcolor{tblStripe}
            $\times$ & $\textcolor{tblWarm}{\times}$
            & 82.83 & 0.2764 & 31.42
            & 84.2 & 0.281 & 49.22 \\

            $\checkmark$ & $\textcolor{tblWarm}{\times}$
            & 82.3 & 0.283 & 43.27
            & 78.5 & 0.289 & 57.23 \\

            \rowcolor{tblStripe}
            $\times$ & $\textcolor{tblGood}{\checkmark}$
            & 85.1 & 0.278 & 41.99
            & 81.7 & 0.284 & 54.24 \\

            \rowcolor{tblLilac}
            $\checkmark$ & $\textcolor{tblGood}{\checkmark}$
            & \tblbest{71.4}
            & \tblbest{0.297}
            & \tblbest{48.63}
            & \tblbest{70.1}
            & \tblbest{0.301}
            & \tblbest{59.55} \\

            \bottomrule
        \end{tabular}%
        }
    \end{minipage}

    \par\vspace{1pt}
    \begin{minipage}[t]{0.99\textwidth}
        \centering
        \textbf{(c) PAM design ablations}\\[2pt]
        \scriptsize
        \setlength{\tabcolsep}{2.0pt}
        \resizebox{\linewidth}{!}{%
        \begin{tabular}{l|ccc|l|ccc|c|ccc}
            \toprule

            \multicolumn{4}{c|}{\cellcolor{tblBlue}\textcolor{tblInk}{\textbf{Denoising phase}}}
            & \multicolumn{4}{c|}{\cellcolor{tblTeal}\textcolor{tblGood}{\textbf{Layer range}}}
            & \multicolumn{4}{c}{\cellcolor{tblPeach}\textcolor{tblWarm}{\textbf{Amplification}}} \\

            \cmidrule(lr){1-4}
            \cmidrule(lr){5-8}
            \cmidrule(lr){9-12}

            \cellcolor{tblBlue}Setting
            & \cellcolor{tblBlue}FID$\downarrow$
            & \cellcolor{tblBlue}CS$\uparrow$
            & \cellcolor{tblBlue}Acc.$\uparrow$
            & \cellcolor{tblTeal}Setting
            & \cellcolor{tblTeal}FID$\downarrow$
            & \cellcolor{tblTeal}CS$\uparrow$
            & \cellcolor{tblTeal}Acc.$\uparrow$
            & \cellcolor{tblPeach}$\alpha$
            & \cellcolor{tblPeach}FID$\downarrow$
            & \cellcolor{tblPeach}CS$\uparrow$
            & \cellcolor{tblPeach}Acc.$\uparrow$ \\

            \midrule

            \rowcolor{tblStripe}
            Early
            & 74.00 & 0.3195 & 64.30
            & Early half
            & 72.40 & 0.3203 & 63.60
            & 1.0
            & 72.03 & 0.3168 & 61.71 \\

            \cellcolor{tblLilac}Middle
            & \cellcolor{tblLilac}\tblbest{68.70}
            & \cellcolor{tblLilac}\tblbest{0.3271}
            & \cellcolor{tblLilac}\tblbest{69.93}
            & \cellcolor{tblLilac}Middle half
            & \cellcolor{tblLilac}\tblbest{68.70}
            & \cellcolor{tblLilac}\tblbest{0.3271}
            & \cellcolor{tblLilac}\tblbest{69.93}
            & 1.1
            & 69.90 & 0.3226 & 66.40 \\

            Late
            & 71.60 & 0.3192 & 62.80
            & Late half
            & 70.90 & 0.3216 & 64.80
            & \cellcolor{tblLilac}1.2
            & \cellcolor{tblLilac}\tblbest{68.70}
            & \cellcolor{tblLilac}\tblbest{0.3271}
            & \cellcolor{tblLilac}\textbf{69.93} \\

            \rowcolor{tblStripe}
            All
            & 71.30 & 0.3251 & 68.10
            & All layers
            & 70.60 & 0.3249 & 67.20
            & 1.4
            & 72.50 & 0.3238 & 67.90 \\

            \bottomrule
        \end{tabular}%
        }
    \end{minipage}
    \par\vspace{-0.5em}
\end{table*}

\vspace{-0.2em}
\subsection{DiT Backbone Transfer}
\label{sec:backbone-generalization}

We evaluate the diffusion-side mechanisms within DeepFusion using PixArt-$\Sigma$~\citep{chen2024pixartsigma} and SD3~\citep{esser2024scaling} as alternative DiT backbones, with the same StyledTextSynth evaluation subset and protocol. For each backbone, we train models with and without text-region weighting and evaluate both checkpoints with and without PAM. Both mechanisms improve CLIPScore and OCR accuracy across the two architectures, while their combination achieves the best result on all three metrics, reaching OCR accuracies of 48.63 on PixArt-$\Sigma$ and 59.55 on SD3 (Table~\ref{tab:diffusion-side-analysis}(b)). These consistent gains show that text-region weighting and PAM transfer across distinct DiT architectures.

\vspace{-0.8em}
\section{Conclusion}
\label{sec:conclusion}

We presented \method, an autonomous visual text generator that makes spatial plans executable through joint autoregressive--diffusion learning. DeepFusion aligns planner states with continuous rendering through planning, spatial, and text-region supervision; PAM strengthens region-specific execution at inference. Experiments show improved planning--rendering consistency and image quality, with a 2B planner and 4B DiT approaching larger models in text accuracy. These findings support joint learning as a basis for faithful plan execution in visual text generation.

\bibliography{iclr2027_conference}
\bibliographystyle{iclr2027_conference}

\clearpage
\appendix
\section{DiT Architecture and Generic Pretraining}
\label{sec:appendix-dit-pretraining}

\paragraph{Architecture.}
Our renderer is a single-stream diffusion transformer with approximately 4B parameters, pretrained from random initialization. The mapper projects the planner states to the DiT width. Noised-image and condition tokens form the joint sequence $[\mathbf Z_t;\mathbf C]$ in Eq.~\ref{eq:joint-sequence}, which is processed by shared attention and feed-forward layers throughout the network. We use the pretrained VAE released with Z-Image~\citep{zimage2025} for image encoding and decoding and keep its parameters frozen. During training, noise is added to the encoded image latents; the DiT operates on patch tokens of these noised latents. At inference, generation starts from latent noise, and the VAE decoder maps the denoised latents to an image. The stated parameter count refers to the DiT alone.

\paragraph{Image-prior initialization.}
We first train the DiT for 200K steps on ImageNet-1K~\citep{deng2009imagenet} to learn an initial image prior before generic text-to-image pretraining.

\paragraph{Generic text-to-image pretraining.}
We pretrain the DiT on a fixed mixture of 15M image--caption pairs at $512\times512$ resolution. The mixture is constructed from Conceptual 12M (CC12M)~\citep{changpinyo2021cc12m}, high-quality subsets of LAION~\citep{schuhmann2022laion5b} and COYO~\citep{kakaobrain2022coyo700m}, RedCaps~\citep{desai2021redcaps}, JourneyDB~\citep{sun2023journeydb}, and WIT~\citep{srinivasan2021wit}. Table~\ref{tab:appendix-pretraining-mixture} reports the exact composition. This stage provides a general image--text prior before the joint DeepFusion training described in Section~\ref{sec:joint-training}.

\paragraph{Optimization.}
For generic text-to-image pretraining, we train on 16 NVIDIA A800 GPUs with a global batch size of 256 for 240K optimizer steps. We use AdamW with a peak learning rate of $1\times10^{-4}$ and weight decay of 0.01. The learning rate is linearly warmed up over the first 7.2K steps (3\% of training) and then decayed with a cosine schedule. The joint stage subsequently optimizes the planner, mapper, DiT, and auxiliary head with the objective in Eq.~\ref{eq:full-objective}; PAM is applied only during inference. Section~\ref{sec:deepfusion-analysis} evaluates the proposed components through controlled comparisons.

\begin{table}[!h]
    \centering
    \small
    \caption{Composition of the 15M-pair generic text-to-image pretraining mixture. The high-quality LAION and COYO subsets are sampled before pretraining.}
    \label{tab:appendix-pretraining-mixture}
    \setlength{\tabcolsep}{4.5pt}
    \resizebox{\linewidth}{!}{%
    \begin{tabular}{lrrl}
        \toprule
        \rowcolor{tblBlue}
        \textcolor{tblInk}{\textbf{Source}} & \textcolor{tblInk}{\textbf{Pairs}} & \textcolor{tblInk}{\textbf{Share}} & \textcolor{tblInk}{\textbf{Primary coverage}} \\
        \midrule
        CC12M & 4.0M & 26.7\% & Real-world images and long-tail concepts \\
        \rowcolor{tblStripe}
        LAION (high-quality subset) & 3.5M & 23.3\% & Visual concepts, styles, and scenes \\
        COYO (high-quality subset) & 2.5M & 16.7\% & Web scenes and contemporary imagery \\
        \rowcolor{tblStripe}
        RedCaps & 2.0M & 13.3\% & Natural user-authored descriptions \\
        JourneyDB & 2.0M & 13.3\% & Aesthetic synthetic images and detailed prompts \\
        \rowcolor{tblStripe}
        WIT & 1.0M & 6.7\% & Entity, cultural, and limited multilingual coverage \\
        \midrule
        \rowcolor{tblLilac}
        \textbf{Total} & \textbf{15.0M} & \textbf{100.0\%} & \\
        \bottomrule
    \end{tabular}%
    }
\end{table}

\section{Human Evaluation and General Generation}
\label{sec:appendix-evaluation}

\subsection{Human Evaluation}

Automatic metrics do not fully capture whether rendered text is visually integrated with its surrounding scene. We therefore conduct a double-blind human evaluation against AnyText~\citep{tuo2023anytext}, TextDiffuser-2~\citep{chen2024textdiffuser}, and FLUX.1 dev~\citep{flux}. We randomly sample 50 prompts from CVTG-2K without cherry-picking and recruit 20 participants, including graphic designers and general users. Model identities are anonymized. Participants score each image from 1 to 10 for visual aesthetics, text quality, and spatial harmony.

As shown in Table~\ref{tab:appendix-user-study}, \method receives the highest scores in text quality and spatial harmony. These results complement the automatic evaluation by measuring whether the planned text is both readable and compositionally compatible with the generated scene.

\begin{table}[!htbp]
    \centering
    \small
    \caption{Human evaluation on 50 randomly sampled CVTG-2K prompts. Scores range from 1 to 10; higher is better.}
    \label{tab:appendix-user-study}
    \setlength{\tabcolsep}{4.5pt}
    \begin{tabular}{l|ccc}
        \toprule
        \rowcolor{tblBlue}
        \textcolor{tblInk}{\textbf{Method}} & \textcolor{tblInk}{\textbf{Vis. Aes.}$\uparrow$} & \cellcolor{tblTeal}\textcolor{tblGood}{\textbf{Text Qual.}$\uparrow$} & \cellcolor{tblPeach}\textcolor{tblWarm}{\textbf{Spatial Harm.}$\uparrow$} \\
        \midrule
        AnyText & 7.12 & 7.85 & 7.43 \\
        \rowcolor{tblStripe}
        TextDiffuser-2 & 7.45 & 8.21 & 7.92 \\
        FLUX.1 dev & \tblbest{8.92} & 8.64 & 8.15 \\
        \midrule
        \rowcolor{tblLilac}
        \method & \tblsecond{8.75} & \tblbest{9.12} & \tblbest{8.95} \\
        \bottomrule
    \end{tabular}
\end{table}

\subsection{General Image Generation}

We additionally evaluate general text-to-image generation after joint training. Table~\ref{tab:appendix-geneval} reports the GenEval~\citep{ghosh2023geneval} results across single-object, two-object, counting, color, position, and color-attribution tasks, with an overall score of 0.91. These results characterize the final model; a comparison with the pretraining checkpoint would be needed to measure changes in general generation capability during joint training.

\begin{table}[!htbp]
    \centering
    \small
    \caption{GenEval~\citep{ghosh2023geneval} results. Sng. Obj.: single object; Tw. Obj.: two objects; Cntng: counting; Clrs: colors; Pstn: position; Clr Attr.: color attribution.}
    \label{tab:appendix-geneval}
    \resizebox{\linewidth}{!}{%
    \begin{tabular}{lccccccc}
        \toprule
        \rowcolor{tblBlue}
        \textcolor{tblInk}{Model} & Sng. Obj. & Tw. Obj. & Cntng & \cellcolor{tblTeal}Clrs & \cellcolor{tblTeal}Pstn & \cellcolor{tblTeal}Clr Attr. & \cellcolor{tblPeach}\textcolor{tblWarm}{Overall} \\
        \midrule
        \rowcolor{tblLilac}
        \method & 0.99 & 0.95 & 0.88 & 0.90 & 0.92 & 0.79 & \tblbest{0.91} \\
        \bottomrule
    \end{tabular}%
    }
\end{table}

\section{Text-Region-Weighted Flow-Matching Loss}
\label{sec:appendix-flow-loss}

Algorithm~\ref{alg:flow-matching-region-loss} details the minibatch computation of the text-region-weighted objective in Eq.~\ref{eq:diffusion-loss}. For each training image, we rasterize its normalized reference boxes onto the latent grid and assign weight $\gamma$ to every covered patch while leaving all other patches at unit weight. The resulting map reweights the squared flow-matching residual, whose target velocity is $\mathbf v=\boldsymbol{\epsilon}-\mathbf z_0$. Overlapping boxes do not compound the weight: a patch covered by one or more boxes always receives weight $\gamma$.

\begin{algorithm}[!htbp]
\caption{Text-Region-Weighted Flow-Matching Loss}
\label{alg:flow-matching-region-loss}
\SetAlgoLined
\DontPrintSemicolon
\KwIn{
    Predictions $\mathbf v_\phi\in\mathbb R^{B\times C\times H\times W}$, \\
    target velocities $\mathbf v=\boldsymbol{\epsilon}-\mathbf z_0\in\mathbb R^{B\times C\times H\times W}$, \\
    normalized reference boxes $\{\mathcal B_n\}_{n=1}^{B}$, \\
    text-region weight $\gamma\geq1$
}
\KwOut{Minibatch loss $\widehat{\mathcal L}_{\mathrm{diff}}$}

\For{$n\leftarrow1$ \KwTo $B$}{
    \tcp{Start with the standard flow-matching weight at every latent patch}
    $\mathbf W_n\leftarrow\mathbf 1^{H\times W}$\;
    
    \tcp{Rasterize each valid normalized box onto the latent grid}
    \ForEach{$b=(x_1,y_1,x_2,y_2)\in\mathcal B_n$}{
        \If{$0\leq x_1<x_2\leq1$ and $0\leq y_1<y_2\leq1$}{
            $j_1\leftarrow\lfloor Wx_1\rfloor$, $j_2\leftarrow\min(W,\lceil Wx_2\rceil)$\;
            $i_1\leftarrow\lfloor Hy_1\rfloor$, $i_2\leftarrow\min(H,\lceil Hy_2\rceil)$\;
            
            \tcp{Use a union mask, so overlaps remain weighted by $\gamma$}
            $\mathbf W_n[i_1:i_2,j_1:j_2]\leftarrow\gamma$\;
        }
    }
    
    \tcp{The channelwise residual norm gives one scalar error per patch}
    $\ell_n\leftarrow\sum_{\mathbf u}\mathbf W_n(\mathbf u)
    \left\|\mathbf v_{\phi,n}(\mathbf u)-\mathbf v_n(\mathbf u)\right\|_2^2$\;
}
$\widehat{\mathcal L}_{\mathrm{diff}}\leftarrow\frac{1}{B}\sum_{n=1}^{B}\ell_n$\;
\Return $\widehat{\mathcal L}_{\mathrm{diff}}$
\end{algorithm}

Here, $\mathbf u$ indexes a spatial location on the $H\times W$ latent grid, and the $\ell_2$ norm sums the residual over its $C$ channels. The floor/ceiling conversion includes every latent patch intersected by a reference box. Consequently, setting $\gamma=1$ exactly recovers the standard flow-matching objective, whereas $\gamma>1$ increases the contribution of the union of text regions without removing supervision from the rest of the image.

\section{Counterfactual Bbox-Following Test}
\label{sec:appendix-adversarial-bbox}

We use a controlled intervention to test whether the DiT follows the planner's bbox representation. For a prompt $p$, we first obtain its clean bbox--content plan $s=\{(b_i,c_i)\}_{i=1}^{K}$ and its condition tokens $\mathbf C$. We then construct a counterfactual plan $\tilde{s}=\{(\tilde b_i,c_i)\}_{i=1}^{K}$ by changing only the box locations. Each $\tilde b_i$ preserves the width and height of $b_i$, remains fully inside the image, and does not overlap another counterfactual box. Candidate locations that violate these validity constraints are rejected. The relative target displacement is $r=\frac{1}{K}\sum_i\frac{\|\operatorname{ctr}(\tilde b_i)-\operatorname{ctr}(b_i)\|_2}{\sqrt{w_i^2+h_i^2}}=0.35$, measured in units of the original bbox diagonal. We run the planner with forced decoding on $\tilde{s}$ and replace only the coordinate-span condition tokens in $\mathbf C$ with their forced-plan counterparts. The prompt condition, text-content condition tokens, DiT weights, and sampling seed remain fixed.

The test directly measures whether the rendered text regions follow the injected target boxes. Let $\hat b_i$ be the OCR-detected box in the counterfactual generation. \emph{Target IoU} is the mean $\operatorname{IoU}(\hat b_i,\tilde b_i)$. We define \emph{center error} relative to the diagonal of the target box,
\begin{equation}
E_{\mathrm{ctr}}=\frac{1}{K}\sum_{i=1}^{K}
\frac{\left\|\operatorname{ctr}(\hat b_i)-\operatorname{ctr}(\tilde b_i)\right\|_2}
{\sqrt{\tilde w_i^2+\tilde h_i^2}},
\label{eq:appendix-center-error}
\end{equation}
where $\operatorname{ctr}(b)$ denotes the box center and $(\tilde w_i,\tilde h_i)$ are the width and height of $\tilde b_i$. Thus, $E_{\mathrm{ctr}}=0$ is exact location following, while $E_{\mathrm{ctr}}=1$ means that the rendered center is one target-box diagonal away. We use the clean-plan center error as the internal reference: a forced-shift error close to the clean value, together with high Target IoU, shows that the DiT executes the counterfactual bbox with comparable spatial precision. Word Accuracy and CLIPScore are reported only to verify that this spatial intervention does not conflate location following with text or global-image degradation. The clean row reuses the full-objective baseline on the four-region CVTG-2K subset in Table~\ref{tab:aux-generation-bridge}: Word Acc.=0.8309, Target IoU=PRC=0.9176, and CLIPScore=0.7925. Center error is recomputed on the same examples for the final comparison.

\begin{table*}[!t]
    \centering
    \scriptsize
    \caption{Counterfactual bbox-following results on the four-region CVTG-2K subset. The forced-shift plan preserves box size, stays in bounds, and contains no overlapping pair; only coordinate-span condition tokens differ from the clean run.}
    \label{tab:appendix-adversarial-protocol}
    \setlength{\tabcolsep}{5pt}
    \resizebox{\textwidth}{!}{%
    \begin{tabular}{l|ccccc}
        \toprule
        \rowcolor{tblBlue}
        \textcolor{tblInk}{\textbf{Condition}} & \cellcolor{tblPeach}\textcolor{tblWarm}{\textbf{Target shift $r$}} & \cellcolor{tblTeal}\textcolor{tblGood}{\textbf{Target IoU}$\uparrow$} & \cellcolor{tblTeal}\textcolor{tblGood}{\textbf{Center Error}$\downarrow$} & \textcolor{tblInk}{\textbf{Word Acc.}$\uparrow$} & \textcolor{tblInk}{\textbf{CLIPScore}$\uparrow$} \\
        \midrule
        \rowcolor{tblStripe}
        Clean self-generated plan & 0.00 & 0.9176 & 0.068 & 0.8309 & 0.7925 \\
        \rowcolor{tblLilac}
        Forced valid-shift plan & 0.35 & 0.9042 & 0.074 & 0.8295 & 0.7941 \\
        \bottomrule
    \end{tabular}%
    }
\end{table*}

At a mean relative target displacement of $r=0.35$, Target IoU decreases by only 0.0134 (0.9176 to 0.9042), while center error changes from 0.068 to 0.074. Word accuracy and CLIPScore remain effectively unchanged. These results reveal \emph{counterfactual spatial controllability}: coordinate-span states are not merely linearly decodable spatial representations, but an executable spatial interface that the DiT follows even when the requested locations depart from the planner's self-generated layout.

\section{Additional Qualitative Results}
\label{sec:appendix-qualitative}

\begin{figure*}[!htbp]
    \centering
    \textbf{(a) Poster and comic design}\\[-0.3em]
    \includegraphics[width=0.95\textwidth]{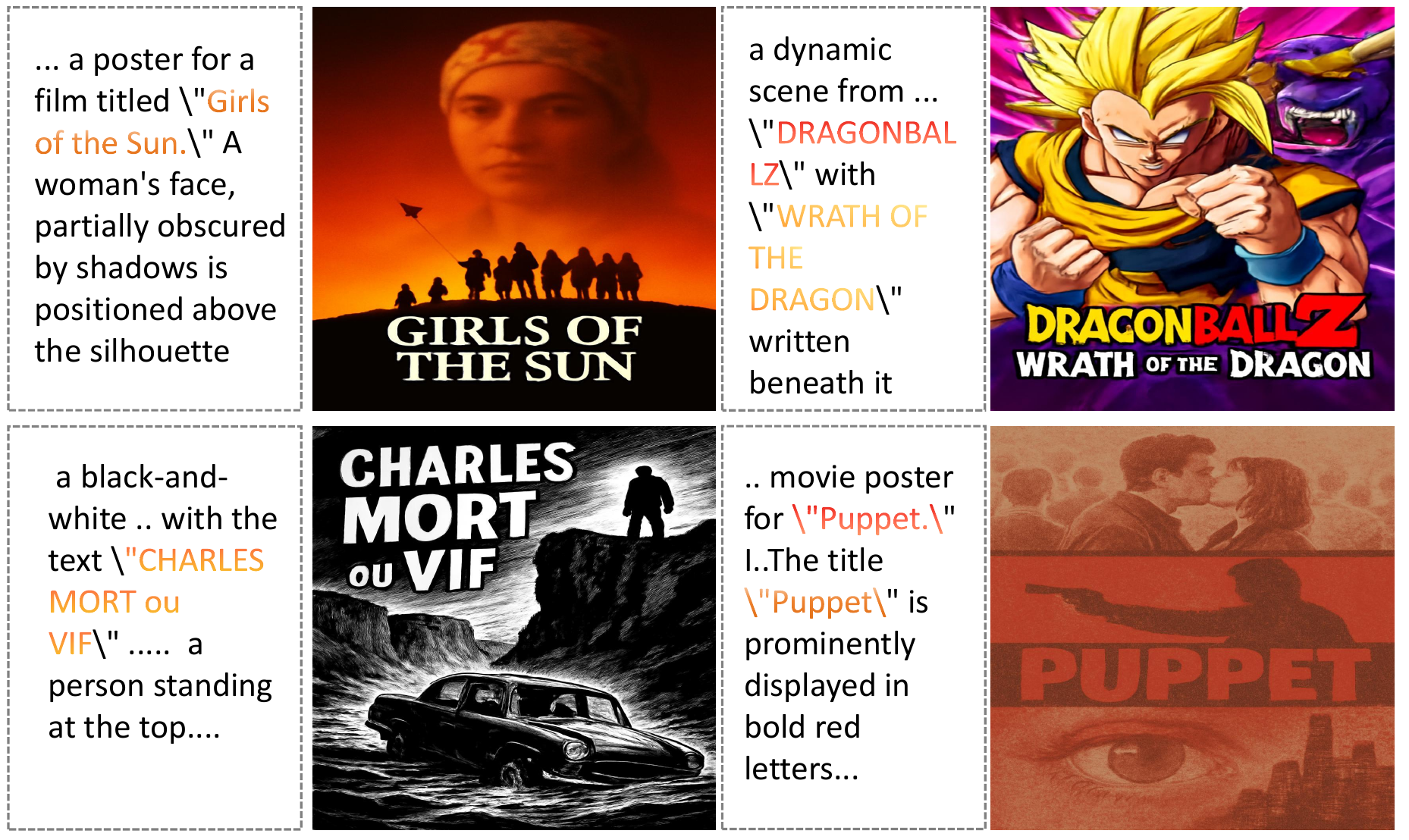}\\[0.7em]
    \textbf{(b) Book-cover and cinematic design}\\[-0.3em]
    \includegraphics[width=0.95\textwidth]{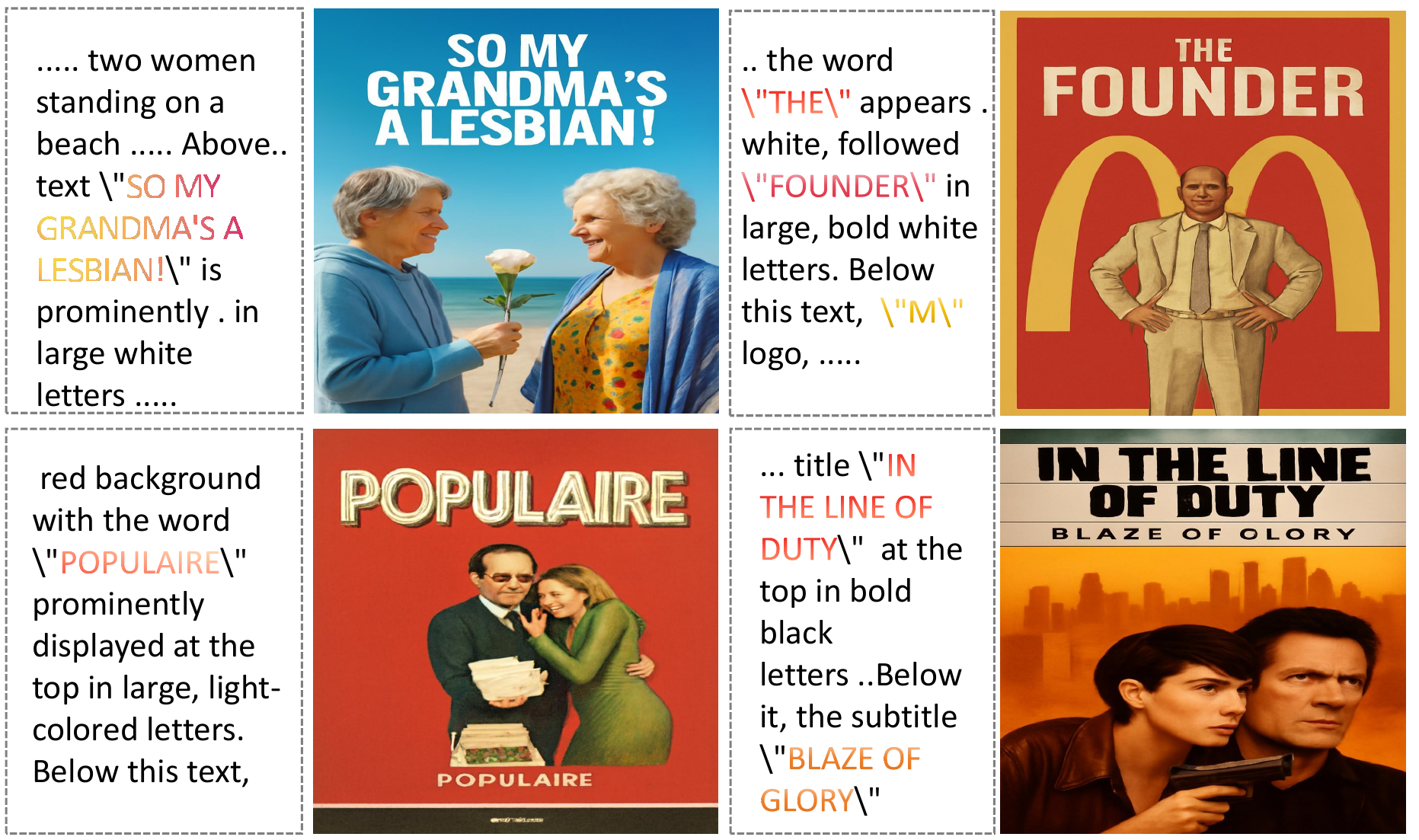}
    \caption{Additional qualitative results in diverse design settings. Panel (a) shows poster and comic examples; panel (b) shows book-cover and cinematic examples.}
    \label{fig:appendix-applications}
\end{figure*}

Figure~\ref{fig:appendix-applications} presents additional examples beyond the benchmark prompts in the main paper. The two panels cover poster/comic and book-cover/cinematic-design settings, where the model must jointly select readable text placement and match the typography to the visual style. These results complement the controlled comparisons in Figure~\ref{fig:qualitative-main} with visually diverse applications.

\section{Theoretical Analysis of Text-Region-Weighted Learning}
\label{sec:appendix-region-theory}

Text-region weighting allocates additional supervision to the spatial support of rendered text while retaining supervision over the complete image. We analyze this design at three levels: the population regression target, the allocation of approximation error, and the local optimization of the joint objective. The results concern the velocity-prediction objective in Eq.~\ref{eq:diffusion-loss}; their assumptions and proofs are given explicitly below.

\paragraph{Setup.}
Let $\vartheta$ collect the trainable parameters, and let $M(\mathbf u)$ indicate membership in the union of the reference text boxes on the latent grid. The mask is computed from training annotations and is independent of $\vartheta$. With $\ell_\vartheta(\mathbf u)=\|v_\vartheta(\mathbf u)-v(\mathbf u)\|_2^2$, define
\begin{equation}
\begin{aligned}
 A(\vartheta)&=\mathbb E\!\left[\sum_{\mathbf u}M(\mathbf u)\ell_\vartheta(\mathbf u)\right],\\
 B(\vartheta)&=\mathbb E\!\left[\sum_{\mathbf u}(1-M(\mathbf u))\ell_\vartheta(\mathbf u)\right],\\
 \mathcal L_\gamma&=B+\gamma A,\qquad
 J_\gamma=R+\lambda\mathcal L_\gamma,\qquad
 R=\mathcal L_{\mathrm{plan}}+\lambda_{\mathrm{aux}}\mathcal L_{\mathrm{aux}},
 \label{eq:theory-risk-decomposition}
\end{aligned}
\end{equation}
where $\lambda=\lambda_{\mathrm{diff}}>0$ and $\gamma\geq1$. Expectations include the training examples, noise, and sampled times. Thus, $J_\gamma$ is exactly the joint objective in Eq.~\ref{eq:full-objective}. $A$ measures error over box-covered patches, including any background within a box; it is not a glyph-segmentation or OCR loss. We assume finite second moments throughout.

\subsection{Preserving the Conditional Regression Target}

\paragraph{Proposition 1 (Consistency under layout-sufficient conditioning).}
Fix a conditioning representation and write $X=(\mathbf z_t,t,\mathbf C,\mathbf u)$ and $V=v(\mathbf u)$. Suppose $M(\mathbf u)$ is a measurable function of $X$. Over unrestricted square-integrable predictors of $X$, the weighted and unweighted regression objectives have the same unique minimizer almost surely, $f^*(X)=\mathbb E[V\mid X]$. Moreover,
\begin{equation}
 \mathcal L_\gamma(f)-\mathcal L_\gamma(f^*)
 =\mathbb E\!\left[\sum_{\mathbf u}W(\mathbf u)
 \|f(X)-f^*(X)\|_2^2\right],\qquad W=1+(\gamma-1)M.
 \label{eq:theory-weighted-excess}
\end{equation}
\emph{Proof.} Expand $\|f-V\|^2$ around $f^*$. The conditional cross term vanishes because $\mathbb E[V-f^*\mid X]=0$ and $W$ is $X$-measurable. Taking expectations and summing over patches gives Eq.~\ref{eq:theory-weighted-excess}. Since $W\geq1$, equality with the minimum holds exactly when $f=f^*$ almost surely. $\square$

If $\mathcal E_{\mathrm{text}}=\mathbb E\sum_{\mathbf u}M\|f-f^*\|^2$ and $\mathcal E_{\mathrm{bg}}=\mathbb E\sum_{\mathbf u}(1-M)\|f-f^*\|^2$, the excess risk $\Delta_\gamma$ in Eq.~\ref{eq:theory-weighted-excess} satisfies
\begin{equation}
 \Delta_\gamma=\gamma\mathcal E_{\mathrm{text}}+\mathcal E_{\mathrm{bg}},
 \qquad \mathcal E_{\mathrm{text}}\leq\Delta_\gamma/\gamma,
 \qquad \mathcal E_{\mathrm{text}}+\mathcal E_{\mathrm{bg}}\leq\Delta_\gamma.
 \label{eq:theory-error-control}
\end{equation}
Consequently, a fixed excess-risk budget imposes a tighter bound on text-region approximation error while still controlling full-image velocity error. This compares error certificates at a given budget; it does not assume identical achieved budgets across training runs.

\paragraph{Role of the spatial interface.}
The assumption holds when the condition determines the reference layout, as an explicit layout does. For the learned states $\mathbf C$, it requires that the mask remain recoverable from those states; auxiliary coordinate supervision supports this design goal but does not by itself prove exact recoverability. For a general fixed representation, set $p_X=\mathbb E[M\mid X]$, $\mu_X=\mathbb E[V\mid X]$, and $s_X^2=\mathbb E[\|V-\mu_X\|^2\mid X]$. Conditional quadratic minimization and Cauchy--Schwarz give
\begin{equation}
\begin{aligned}
 f_\gamma^*(X)&=\frac{\mathbb E[WV\mid X]}{\mathbb E[W\mid X]},\\
 \|f_\gamma^*(X)-\mu_X\|_2
 &\leq\frac{(\gamma-1)\sqrt{p_X(1-p_X)}\,s_X}
 {1+(\gamma-1)p_X}.
 \label{eq:theory-imperfect-conditioning}
\end{aligned}
\end{equation}
Indeed, the difference equals $(\gamma-1)\mathbb E[(M-p_X)(V-\mu_X)\mid X]/[1+(\gamma-1)p_X]$, and the conditional variance of $M$ is $p_X(1-p_X)$. Thus target invariance is exact for a recoverable mask, while any shift is controlled by conditional mask uncertainty and the weight increment. This connects informative spatial conditions with reliable region weighting. The next two results allow arbitrary learned representations.

\subsection{Prioritizing Text Under Shared Model Capacity}

\paragraph{Proposition 2 (Regional priority at joint optima).}
Keep the parameter space, data distribution, $R$, and $\lambda$ fixed. For $\gamma_2>\gamma_1\geq1$, suppose $\vartheta_i$ globally minimizes $J_{\gamma_i}$. Then
\begin{equation}
 A(\vartheta_2)\leq A(\vartheta_1).
 \label{eq:theory-monotone-risk}
\end{equation}
More generally, if $J_{\gamma_i}(\vartheta_i)\leq\inf_\vartheta J_{\gamma_i}(\vartheta)+\delta_i$ with $\delta_i\geq0$, then
\begin{equation}
 A(\vartheta_2)-A(\vartheta_1)
 \leq\frac{\delta_1+\delta_2}{\lambda(\gamma_2-\gamma_1)}.
 \label{eq:theory-approximate-optima}
\end{equation}
\emph{Proof.} Set $F=R+\lambda B$. Approximate optimality yields
$F(\vartheta_1)+\lambda\gamma_1 A(\vartheta_1)\leq F(\vartheta_2)+\lambda\gamma_1 A(\vartheta_2)+\delta_1$
and the analogous inequality with indices exchanged. Adding cancels $F$ and gives
$\lambda(\gamma_2-\gamma_1)(A(\vartheta_2)-A(\vartheta_1))\leq\delta_1+\delta_2$.
Setting both gaps to zero proves Eq.~\ref{eq:theory-monotone-risk}. $\square$

This establishes regional priority even when planning and rendering share parameters and cannot fit every target perfectly. It does not require convexity, but the exact statement concerns global optima; Eq.~\ref{eq:theory-approximate-optima} exposes the dependence on optimization quality. Background, plan, or auxiliary errors can trade off against $A$, explaining why a moderate weight is appropriate for a joint objective.

\paragraph{Effective spatial allocation and bounded weights.}
For a fixed grid size, let $\rho\in(0,1)$ be the probability that a uniformly sampled patch lies inside a reference box. Under the probability measure obtained by weighting example--patch pairs by $W/\mathbb E[W]$, the text-region probability becomes
\begin{equation}
 q_\gamma=\frac{\gamma\rho}{1+(\gamma-1)\rho},\qquad
 \frac{q_\gamma}{1-q_\gamma}=\gamma\frac{\rho}{1-\rho},\qquad
 \mathcal L_1\leq\mathcal L_\gamma\leq\gamma\mathcal L_1.
 \label{eq:theory-allocation}
\end{equation}
The first two identities follow by normalizing the text and background masses $\gamma\rho$ and $1-\rho$; the last follows pointwise from $1\leq W\leq\gamma$. This is an interpretation of the existing objective, not per-image loss normalization. It shows that weighting increases the relative emphasis on text rather than uniformly rescaling the loss. The union-mask construction in Algorithm~\ref{alg:flow-matching-region-loss} keeps these bounds independent of the number of overlapping boxes, and every background patch retains a positive loss coefficient.

\subsection{A Local Optimization Advantage for Text Regions}

\paragraph{Proposition 3 (Finite-step regional descent advantage).}
At a common parameter point $\vartheta$, set $a=\nabla A(\vartheta)$, $g=\nabla J_1(\vartheta)$, and $c=\lambda(\gamma-1)$. Assume $A$ has an $L_A$-Lipschitz gradient on a convex neighborhood containing the two updates
$\vartheta_1^+=\vartheta-\eta g$ and $\vartheta_\gamma^+=\vartheta-\eta(g+ca)$, where $\eta>0$. Then
\begin{equation}
 A(\vartheta_\gamma^+)-A(\vartheta_1^+)
 \leq-\eta c\|a\|_2^2
 +\frac{L_A\eta^2}{2}\left(\|g+ca\|_2^2+\|g\|_2^2\right).
 \label{eq:theory-local-descent}
\end{equation}
In particular, for $\gamma>1$, $a\neq0$, and $L_A>0$, the weighted update has strictly smaller text-region risk whenever
\begin{equation}
 0<\eta<\frac{2c\|a\|_2^2}
 {L_A\left(\|g+ca\|_2^2+\|g\|_2^2\right)},
 \label{eq:theory-step-condition}
\end{equation}
provided both updates remain in the neighborhood.
\emph{Proof.} Equation~\ref{eq:theory-risk-decomposition} gives $\nabla J_\gamma=g+ca$. Apply the smoothness upper bound to $A(\vartheta_\gamma^+)$ and the corresponding lower bound to $A(\vartheta_1^+)$. Subtracting cancels $A(\vartheta)$ and the common first-order term $-\eta\langle a,g\rangle$, leaving Eq.~\ref{eq:theory-local-descent}. Equation~\ref{eq:theory-step-condition} makes its right-hand side negative. If $L_A=0$, the remainder vanishes directly. $\square$

Thus, at a fixed parameter point, region weighting contributes an additional first-order decrease $\eta\lambda(\gamma-1)\|\nabla A\|^2$ relative to the unweighted joint update. Actual descent from $A(\vartheta)$ additionally requires $\langle a,g\rangle+c\|a\|^2>0$ and a sufficiently small step. This makes the mechanism explicit: the added regional term can offset a conflicting contribution from the other objectives. Proposition 3 describes full-gradient descent locally; it does not assert a global convergence-rate ordering for stochastic AdamW.

\paragraph{Implications for DeepFusion.}
Together, these results justify text-region weighting as a bounded allocation of learning emphasis: a layout-sufficient condition preserves the population target, the joint objective prioritizes text-region error under shared capacity, and the added gradient term favors local regional improvement. These properties complement the spatial interface learned with auxiliary supervision and provide an analytical basis for the region-specific execution design. The OCR and FID improvements in Table~\ref{tab:diffusion-side-analysis} provide the corresponding empirical assessment, since velocity-error guarantees alone do not imply monotonic improvements in perceptual or recognition metrics.

\section{Limitations}
\label{sec:appendix-limitations}

\method is trained and evaluated primarily on English-centric text-image data. Preliminary qualitative evaluation on Korean and Japanese prompts suggests that its performance does not yet transfer reliably to non-Latin scripts. While the model often preserves the intended composition and approximates the visual density of the requested script, generated glyphs can exhibit inaccurate stroke structures or character identities. This observation is consistent with the limited coverage of non-Latin scripts in the current training distribution. Extending the framework with large-scale multilingual text-image data and script-aware supervision represents an important direction for future work.

\begin{figure}[!htbp]
    \centering
    \includegraphics[width=\linewidth]{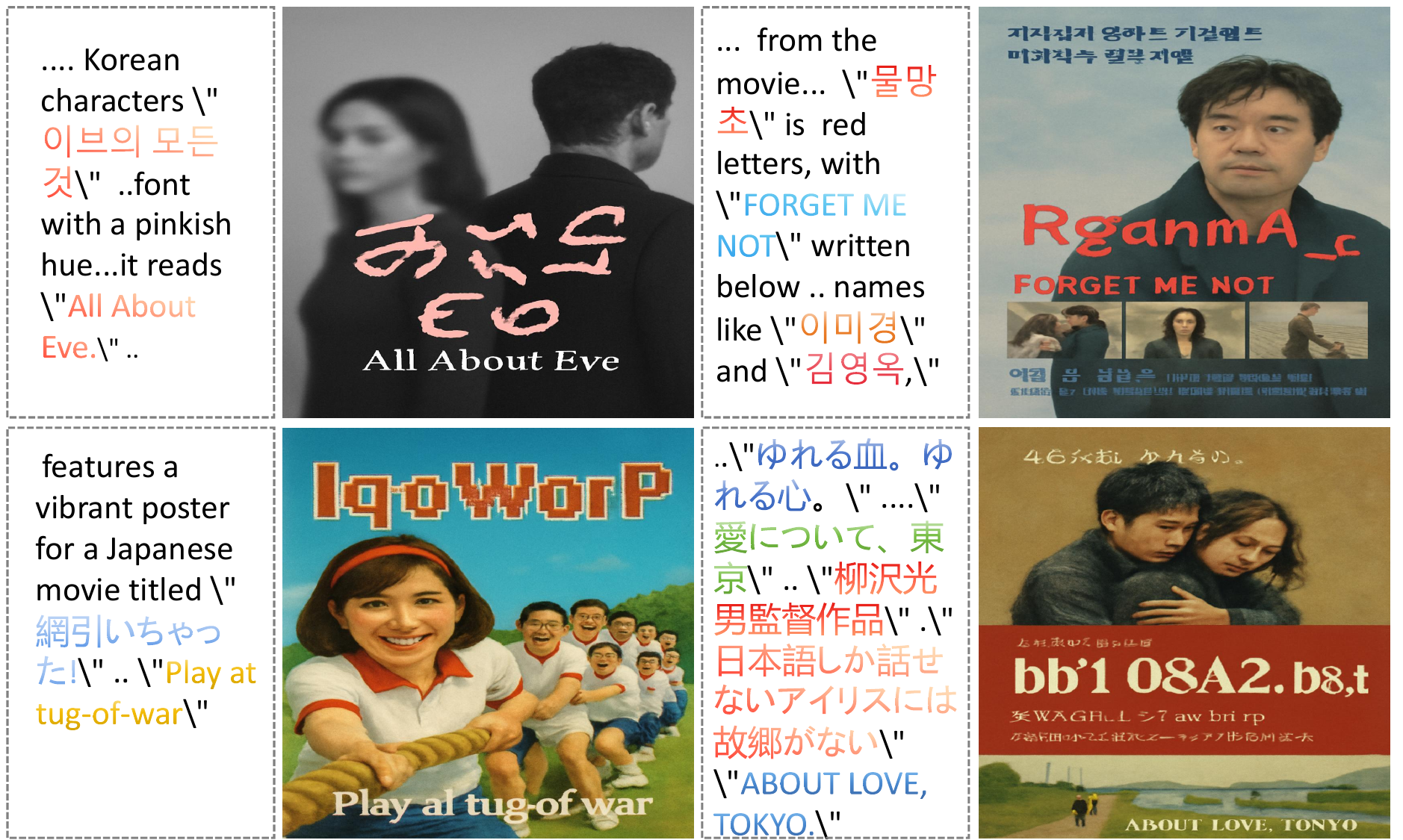}
    \caption{Qualitative examples on Korean (top) and Japanese (bottom) prompts, which are underrepresented in the current training distribution. \method preserves the overall composition and script-like visual appearance, but may produce inaccurate strokes or character identities.}
    \label{fig:appendix-multilingual-failures}
\end{figure}

\end{document}